\pdfoutput=1

\documentclass[11pt]{article}

\usepackage[final]{acl}

\usepackage{times}
\usepackage{latexsym}
\usepackage{comment}
\usepackage{multirow}
\usepackage{tabulary}
\usepackage{tabularx}
\usepackage{amsmath}
\usepackage{amsfonts} 
\usepackage[T1]{fontenc}

\usepackage[utf8]{inputenc}

\usepackage{microtype}

\usepackage{inconsolata}

\usepackage{array}
\usepackage{graphicx}
\usepackage{tabulary}
\usepackage{colortbl}
\usepackage{booktabs}
\usepackage{enumitem}
\usepackage{makecell}
\title{Learning to Summarize from LLM-generated Feedback}

\author{Hwanjun Song$^{1}$\thanks{~~Corresponding Author.},\\~\textbf{Taewon Yun}$^{1}$,~\textbf{Yuho Lee}$^{1}$,~\textbf{Jihwan Oh}$^{1}$,~\textbf{Gihun Lee}$^{2,}$\thanks{~~This work is conducted independently and is not related to the author(s)' position at Hyperconnect and Amazon.}, \textbf{Jason Cai}$^{3,\dagger}$~,~ \textbf{Hang Su}$^{3,\dagger}$\\
$^{1}$Korea Advanced Institute of Science and Technology\\ $^{2}$Hyperconnect\\ $^{3}$AWS AI Labs\\
\ songhwanjun@kaist.ac.kr}

\definecolor{blue0}{RGB}{50, 130, 150}  
\definecolor{blue1}{RGB}{70, 130, 180}\colorlet{blue1}{blue0!70}  
\definecolor{blue2}{RGB}{70, 130, 180}\colorlet{blue2}{blue0!50}  
\definecolor{blue3}{RGB}{70, 130, 180}\colorlet{blue3}{blue0!30}  
\definecolor{blue4}{RGB}{70, 130, 180}\colorlet{blue4}{blue0!10}

\newcommand{\algname}{{FeedSum}} 
\begin{document}
\maketitle
\begin{abstract}
Developing effective text summarizers remains a challenge due to issues like hallucinations, key information omissions, and verbosity in LLM-generated summaries. This work explores using LLM-generated feedback to improve summary quality by aligning the summaries with human preferences for faithfulness, completeness, and conciseness. We introduce FeedSum, a large-scale dataset containing multi-dimensional LLM feedback on summaries of varying quality across diverse domains. Our experiments show how feedback quality, dimensionality, and granularity influence preference learning, revealing that high-quality, multi-dimensional, fine-grained feedback significantly improves summary generation. We also compare two methods for using this feedback: supervised fine-tuning and direct preference optimization. Finally, we introduce SummLlama3-8b, a model that outperforms the nearly 10x larger Llama3-70b-instruct in generating human-preferred summaries, demonstrating that smaller models can achieve superior performance with appropriate training.
The full dataset and SummLlama3-8B model are available at \url{https://huggingface.co/datasets/DISLab/FeedSum} and \url{https://huggingface.co/DISLab/SummLlama3-8B}. 
\end{abstract}

\section{Introduction}
\label{sec:introduction}

Developing an effective text summarizer has long been a challenge, as summaries generated by language models often fall short of human standards \cite{maynez2020faithfulness, el2021automatic, roit2023factually}. While large language models (LLMs) have greatly improved the coherence and fluency of summaries \cite{liu2023g}, persistent issues remain, such as unfaithful statements (hallucinations), omission of key information (low completeness), and verbosity (low conciseness) in the summaries \cite{lee2023unisumeval, song2024finesure}.

An ideal approach would involve providing expert-level summary examples for language models to imitate. However, creating such high-quality summaries is both labor-intensive and difficult to scale effectively.
A better alternative is to leverage feedback on the summaries through reinforcement learning from human feedback (RLHF) \cite{stiennon2020learning,  rafailov2024direct}. RLHF avoids the need to write an ideal summary by having users select their preferred response from candidate summaries of the same document. 
Yet, human involvement still poses scalability challenges, particularly when acquiring high-quality, fine-grained human feedback across {multiple} dimensions, such as faithfulness and completeness. For example, \citet{lee2023unisumeval} reports that the cost of obtaining fine-grained human feedback for these three dimensions exceeded \$30K for just 2,025 summaries.

\begin{figure}[t!]
\begin{center}
\includegraphics[width=7.7cm]{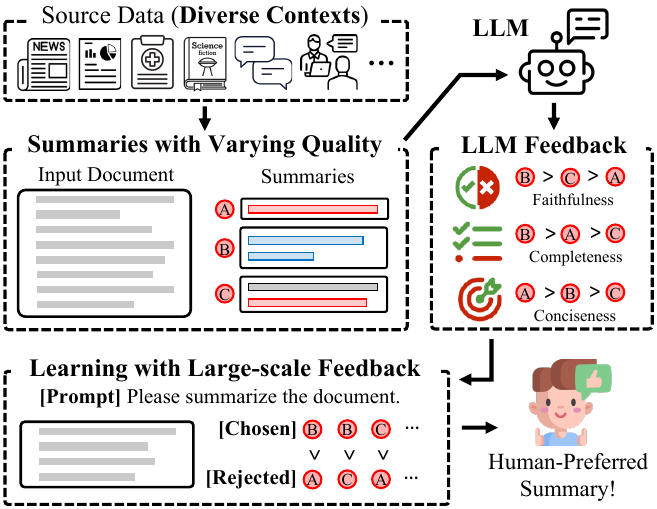}
\end{center}
\vspace*{-0.35cm}
\caption{\algname{} consists of summaries of varying quality, generated by 13 different summarizers across input documents from 7 distinct domains. Through automated evaluation using LLMs, 125K document-summary pairs have been produced, each accompanied by LLM-generated multi-dimensional feedback, providing valuable data for preference learning.}
\label{fig:overview}
\vspace*{-0.4cm}
\end{figure}

In this work, we address this challenge by utilizing \emph{LLM-generated} feedback, known as RL from AI feedback (RLAIF) \cite{lee2023rlaif, dutta2024applying}, with a special focus on text summarization.
Our approach shifts focus to the relatively unexplored area of leveraging LLM-generated feedback to enhance summary quality, whereas most existing research in summarization has primarily concentrated on using LLMs to evaluate summaries. \cite{wan2024acueval,tang2024minicheck, song2024finesure}. 
Specifically, our goal is to produce human-preferred summaries by exploiting LLM feedback with respect to the three core dimensions, namely \emph{faithfulness}, ensuring summaries are consistent with original documents; \emph{completeness}, encompassing all key-facts\footnote{A key-fact is a concise sentence that conveys a single key piece of information \cite{bhandari2020re, song2024finesure}.}; and \emph{conciseness}, maintaining a succinct and focused summary. We focus on these three dimensions, as LLMs already excel in other axes like fluency and coherence \cite{liu2023g, song2024finesure}. Figure \ref{fig:overview} illustrates our complete pipeline for learning from LLM-generated feedback, consisting of data sourcing, feedback generation, and preference learning.

We begin by creating a large-scale dataset, \textbf{\algname{}}, with {multi-dimensional} LLM feedback on text summaries. To obtain summaries with varying levels of quality, we generate them using 13 different language models, including 3 non-LLMs ({e.g.}, Bart), 7 open-source LLMs ({e.g.}, Llama3), and 3 proprietary LLMs ({e.g.}, GPT-4-turbo). For summary generation, these models are applied to a diverse range of input documents, spanning both short and lengthy texts, including dialogue and non-dialogue formats, and across 7 distinct domains. 

The effectiveness of LLM feedback on preference learning can vary based on three factors, including the \emph{quality} of the feedback (e.g., low vs. high), its \emph{dimensionality} (e.g., single vs. multiple), and its level of \emph{granularity} in scoring (e.g., coarse vs. fine). To explore the impact of these factors, we configure four different setups (see Table \ref{table:exp_configuration} for details). Each setup generates 125K pairs of input documents and summaries, accompanied by distinct LLM-generated feedback responses. 
In particular, we find that these factors significantly affect the distribution of feedback scores, which is crucial for determining "chosen" versus "rejected" summaries in preference learning.

Through extensive experiments with FeedSum, we provide insights into learning to summarize from LLM feedback, focusing on three key open questions: \emph{(Q1)} The impact of these three factors on the effectiveness of preference learning; \emph{(Q2)} An analysis on the effectiveness of each dimension in multi-dimensional feedback; and \emph{(Q3)} A comparison between two approaches for utilizing LLM-generated feedback: supervised fine-tuning (SFT) and direct preference optimization (DPO).

Our main contributions are: (1) We create and release \algname{}, the first large-scale summarization dataset for preference learning, featuring high diversity in inputs, summaries, and feedback; (2) We examine how different configurations of LLM-generated feedback impact preference learning, and the importance of feedback quality, dimensionality, and granularity. (3) We examine the alignment trade-off associated with relying on a single dimension for preference learning. (4) We compare the effectiveness of DPO and SFT variants. (5) We release SummLlama3-8b, which outperforms the nearly 10x larger Llama3-70b-instruct in the three human-preferred dimensions.

\vspace*{-0.00cm}
\section{Related Work}
\label{sec:related-work}
\vspace*{-0.1cm}

\newcolumntype{L}[1]{>{\raggedright\let\newline\\\arraybackslash\hspace{0pt}}m{#1}}
\newcolumntype{X}[1]{>{\centering\let\newline\\\arraybackslash\hspace{0pt}}p{#1}}

\begin{table*}[t]
\begin{center}
\scriptsize
\begin{tabular}{|L{1.7cm} |X{1.4cm} X{1.6cm} X{1.2cm} X{1.6cm} |X{2.0cm} X{1.2cm} X{1.8cm}| }\toprule
\!Dataset & \# of Domain & Input Type & \!\!Input Length\!\! & \!\!\!\!Summary Gen.\!\!\!\! & \# of Feedback Dim. & \!\!\!Feedback Size\!\!\! & \!Feedback Type\! \\ \midrule
\!{UniSumEval} & Multiple (9) & \!\!Dial. \& Non-dial.\!\! & \!\!Short--Long\!\! & \!\!Summarizers (9)\!\! & Faith, Comp, Cons & 1K & Human-annotated  \\ \midrule
\!{SynFacEdit} & Single (1) & Non-dial. & Short & \!\!Synthetic Edit (1)\!\! & Faith  & 5K & LLM-generated \\ 
\!{{FeedSum}\,(Ours)}\!\!\!\! & Multiple (7) & \!\!Dial. \& Non-dial.\!\! & \!\!Short--Long\!\! & \!\!Summarizers (13)\!\! & Faith, Comp, Cons & 125K & LLM-generated \\ \bottomrule
\end{tabular}
\end{center}
\vspace*{-0.4cm}
\caption{Comparison of \algname{} with existing summarization datasets with human feedback, UniSumEval\,\cite{lee2023unisumeval}, and LLM feedback, SynFacEdit\,\cite{mishra2024synfac}. The numbers in the parenthesis are the number of input domains (in "\# of Domain") and summarizers used for summary generation (in "Summary Gen"). } 
\label{table:data-stat}
\vspace*{-0.4cm}
\end{table*}

\paragraph{Preference Optimization.}

Preference optimization plays a crucial role in bridging the gap between human intent and the outputs generated by LLMs \cite{yang2024preference, jiang2024survey,rafailov2024direct, ryu2024multi}. The predominant methods are PPO \cite{schulman2017proximal}\footnote{PPO is not well-suited for tasks with lengthy input and output, such as text summarization, due to its iterative auto-regressive generation of summaries at each training iteration. Therefore, we primarily focus on using DPO.}, which fits a reward model to optimize LLMs to generate responses that receive high reward, and DPO \cite{rafailov2024direct}, which directly optimizes the LLMs' outputs based on preference data without relying on an explicit reward model.
These methods have demonstrated effectiveness in aligning LLMs with human preferences, particularly in mitigating hallucination, harmful outputs, and biased contents\,\cite{tonmoy2024comprehensive, bai2024hallucination, allam2024biasdpo, li2024preference}. 
%
Despite the success of preference optimization in other domains, in text summarization, limited work has focused on aligning outputs with human preferences. \citet{stiennon2020learning} collected a comparison dataset to train a PPO reward model, using Reddit posts and coarse evaluations of two summaries, without accounting for multi-dimensional aspects of summarization. Recently, \citet{mishra2024synfac} applied DPO for summarization alignment, but focusing solely on faithfulness using low-quality summaries by synthetically altering high-quality ones with LLMs. 



\paragraph{Automated Evaluation.}
Conventional metrics, such as ROUGE and BERTScore, have shown a weak correlation with human judgment in the evaluation of text summaries\,\cite{pagnoni2021understanding, tang2024tofueval, song2024finesure}. In response, several methods have emerged, focusing on fine-tuning models using well-curated datasets, including natural language inference (NLI)-based and question answering (QA)-based approaches\,\cite{fabbri2022qafacteval, laban2022summac, zhong2022towards}. However, they are typically limited to evaluating only the faithfulness of summaries and require training specialized models. 
The emergence of large language models (LLMs) facilitates multi-dimensional evaluation by incorporating them into human-like evaluation pipelines\,\cite{wang2023chatgpt, tang2024minicheck, van2024field, fu2024gptscore}. Particularly, G-Eval \cite{liu2023g} uses GPT-4 for automated evaluation, providing 1-5 Likert-scale scores based on the prompt. FineSurE \cite{song2024finesure} adopts fine-grained protocols for sentence-level faithfulness and key-fact-level completeness and conciseness, producing objective percentage scores, such as the proportion of faithful sentences and key-facts included.

In this work, we integrate automated evaluation into preference learning, enabling large-scale, fine-grained feedback that addresses three key alignment dimensions of text summarization: faithfulness, completeness, and conciseness.

\section{Data and Experiment Details}

\paragraph{High-level Overview.}
We overview the overall pipeline from data creation to preference learning, as in Figure \ref{fig:overview}, following three distinct steps:

\underline{\emph{Step 1.}} \emph{Input sourcing and summary generation:} Input documents are extracted from 7 diverse sources, varying in domain, length, and type. Summaries are then generated using 13 non-LLMs, open-source LLMs, and proprietary LLMs, producing a wide range of summary quality.

\underline{\emph{Step 2.}} \emph{Feedback generation using LLMs:} 
Feedback is generated through LLM-based summary evaluations using four configurations, adjusting the quality (low vs. high), dimensionality (single vs. multi-dimensional), and granularity (coarse- vs. fine-grained) of the feedback.

\underline{\emph{Step 3.}} \emph{Learning with large-scale feedback:} We examine the potential of machine feedback from LLMs through the lens of preference learning.

\subsection{FeedSum: Data with LLM Feedback}
\label{sec:data-creation}

\newcolumntype{L}[1]{>{\raggedright\let\newline\\\arraybackslash\hspace{0pt}}m{#1}}
\newcolumntype{X}[1]{>{\centering\let\newline\\\arraybackslash\hspace{0pt}}p{#1}}

\begin{table*}[t]
\begin{center}
\footnotesize
\begin{tabular}{|X{0.3cm} |X{2.1cm} X{1.3cm} X{2.5cm}  |L{7.7cm}|}\toprule
\!\!ID\!\! & Quality & \!\!Dimension\!\!\!\!\! & Granularity & \,\,\,\,\,\,\,\,\,\,\,\,\,\,\,\,\,\,\,\,\,\,\,\,\,\,\,\,\,\,\,\,\,\,\,\,\,\,\,\,\,\,\,\,\,\,\,\,\,\,Feedback Example \\ \midrule
\!\!C1\!\! &  \!Low (Llama3-8b)\!\!\!\!\! & Single\!\!\!\!  & Summary-level  & \{Overall Score: 4\}  \\ 
\!\!C2\!\! &  \!High (Llama3-70b)\!\!\!\!\! & Single\!\!\!\! & Summary-level & \{Overall Score: 5\} \\
\!\!C3\!\! &  \!High (Llama3-70b)\!\!\!\!\! & Multiple\!\!\!\! & Summary-level & \{Faithfulness: 5, Completeness: 3, Conciseness: 3\} \\
\!\!C4\!\! &  \!High (Llama3-70b)\!\!\!\!\! & Multiple\!\!\!\!  & \!Sentence \& key-fact\! & \{Faithfulness: 75\%, Completeness: 66\%, Conciseness: 25\%\}\!\!\\ \bottomrule
\end{tabular}
\end{center}
\vspace*{-0.4cm}
\caption{Four different configurations to acquire LLM-generated feedback on the quality of text summaries. } 
\label{table:exp_configuration}
\vspace*{-0.4cm}
\end{table*}

Table \ref{table:data-stat} contrasts our {FeedSum} dataset with two existing datasets with either human-annotated or LLM-generated feedback. \algname{} features input text that is as diverse as UniSumEval\,\cite{lee2023unisumeval}, while simultaneously providing a significantly larger amount of LLM-generated feedback compared to SynFacEdit\,\cite{mishra2024synfac}. 

\paragraph{Input Text Sourcing}
\label{sec:input-text}

The diversity of source documents is crucial for acquiring comprehensive feedback, as it helps identify weaknesses in modern text summarizers across various aspects such as input domain, length, and type (dialogue vs. non-dialogue) \cite{lee2023unisumeval}. Hence, we extract input documents from multiple source datasets encompassing 7 different domains, ranging from short to lengthy input texts, and covering both non-dialogue and dialogue formats. We sample 2,000 input texts from the training set of each source dataset, including four non-dialogue datasets -- CNN/DM (news)\,\cite{nallapati2016abstractive}, Wikihow (lifestyle)\,\cite{koupaee2018wikihow}, GovReport (report)\,\cite{huang2021efficient}, PubMed (medical literature)\,\cite{cohan2018discourse} -- and three dialogue datasets -- DialogSum (dailylife)\,\cite{chen2021dialogsum},  MediaSum (interview)\,\cite{zhu2021mediasum}, MeetingBank (meeting)\,\cite{hu2023meetingbank}. As a result, a total of 14K input documents are sampled.

\paragraph{Summaries with Varying Quality.}
\label{sec:generaged-summary}
The performance of summarization can vary depending on the summarizer chosen, as there is no single model that consistently produces the best quality summary \cite{song2024finesure}. From the perspective of preference learning, it is crucial to gather feedback from a wide range of summaries with varying quality. This allows for pairwise comparisons of different quality summaries even for the same document. Therefore, we choose 13 language models grouped into three categories, namely \emph{non-LLMs}, including Bart-large\,\cite{lewis-etal-2020-bart}, Pegasus-large\,\cite{zhang2020pegasus}, and Flan-t5-large\,\cite{chung2024flant5}, \emph{open-source LLMs}, including Phi-2\,\cite{javaheripi2023phi}, Mistral-7b-instruct\,\cite{jiang2023mistral}, Mixtral-8x7b-instruct\,\cite{jiang2024mixtral}, Llama2-13b-chat\,\cite{touvron2023llama2}, Mistral-nemo-instruct, Llama3-70b-instruct, and Gemma2-27b-instruct\,\cite{team2024gemma2}, and \emph{proprietary LLMs}, including Claude-instant, GPT-3.5-turbo, and GPT-4-turbo\,\cite{achiam2023gpt4}. 
These summarizers generate rich, diverse summaries, producing 182K document--summary pairs.

\subsection{LLM-generated Feedback}
\label{sec:llm-feedback}

We generate feedback by conducting automated evaluations using LLMs. Yet, the effectiveness of LLM feedback varies the evaluation configuration employed. This study addresses open questions about leveraging LLM-based feedback for preference learning, with a focus on its "quality," "dimensionality," and "granularity." These aspects are investigated by contrasting the effectiveness of feedback generated from four distinct configurations (C1--C4) in Table \ref{table:exp_configuration}, as summarized below:

\smallskip\smallskip\noindent
$\bullet$ \textbf{{Feedback Quality (C1 vs. C2)}}: The quality of generated feedback plays a pivotal role in preference learning. To assess the importance of feedback quality, we adjust the capacity of the selected LLMs for feedback generation. We use two open-source LLMs of different sizes: Llama3-8b-instruct for low-quality feedback and Llama3-70b-instruct for high-quality feedback, respectively.

\footnotetext[3]{We exclude cases where the document length exceeds the model's capacity (8K tokens by Llama3 on a single NVIDIA H100 GPU), as well as instances of erroneous feedback from the LLMs, such as incomplete or nonsensical responses.}

\smallskip\smallskip\noindent
$\bullet$ \textbf{{Feedback Dimensionality (C2 vs. C3)}}: The simplest way to gather feedback is to assess the quality of the summary with a single score on a 1–5 Likert scale \cite{wang2023chatgpt}. However, it overlooks key multi-dimensional aspects of summary quality, such as faithfulness, completeness, and conciseness\,\cite{lee2023unisumeval}. Therefore, a more advanced approach involves conducting a multi-dimensional evaluation using LLMs across these three dimensions, yielding a separate score for each \cite{zhong2022towards, liu2023g}. 

\smallskip\smallskip\noindent
$\bullet$ \textbf{{Feedback Granularity (C3 vs. C4)}}: 
Coarse-grained evaluation uses a Likert scale (1--5), but these subjective scores often skew toward higher ratings due to a lack of grounding \cite{wang2023chatgpt, liu2023g}. In contrast, fine-grained evaluation assesses at the sentence or key-fact level, measuring faithfulness, completeness, and conciseness through factual sentence proportions and key-fact coverage, yielding percentage scores better correlate with human feedback \cite{song2024finesure}.

\smallskip
Thus, all summary--document pairs are subjected to LLM-based summarization evaluation for each configuration. 125K summary-document pairs remain after excluding failed and special cases.\footnotemark[3] More details including evaluation prompts, methodologies, and data statistics are in Appendix \ref{sec:feedback-generation}.

\subsection{Learning with Large-scale Feedback}
\label{sec:preference-learning}

We primarily focus on three key questions: 

\emph{Q1:} How do the quality, dimensionality, and granularity of LLM-generated feedback (C1--C4 in Table \ref{table:exp_configuration}) influence preference learning? 

\emph{Q2:} What impact does each dimension have in the case of multi-dimensional feedback? 

\emph{Q3:} How much can DPO enhance the quality of summaries compared to SFT variants? 

The experimental setups are detailed below and see Appendix \ref{sec:train-configuration} for the detailed training configuration and input--output format for SFT and DPO.

\begin{figure*}[t!]
\begin{center}
\includegraphics[width=16.1cm]{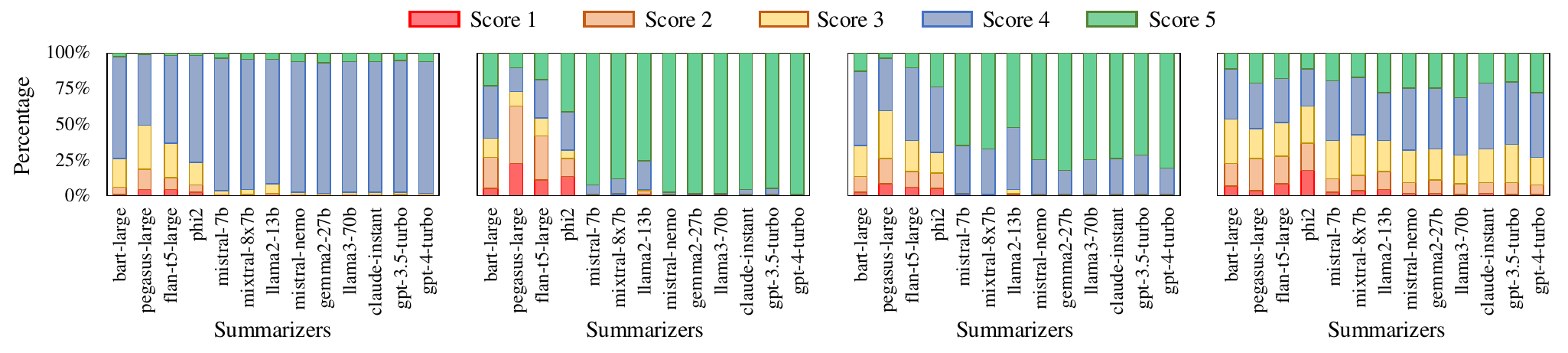}
\end{center}
\vspace*{-0.35cm}
{\small \hspace*{1.2cm} (a) Configuration\,1 (C1). \hspace*{0.5cm} (b) Configuration 2\,(C2). \hspace*{0.6cm} (c) Configuration 3\,(C3). \hspace*{0.55cm} (d) Configuration 4\,(C4).}
\vspace*{-0.25cm}
\caption{Distribution of summary scores on a 1–5 Likert scale across the four different configurations. Percentage scores in C4 are converted into Likert-scale ones through uniform quantization for ease of interpretation.}
\label{fig:feedback-distribution}
\vspace*{-0.4cm}
\end{figure*}

\paragraph{Q1: Impact of Feedback Configuration.}
In this experiment, we consistently train Llama3-8b-instruct using DPO \cite{rafailov2024direct}, but with feedback generated by differently configured LLMs (C1--C4 in Table \ref{table:exp_configuration}). Specifically, for each configuration, we create a set of paired summaries -- one chosen and one rejected -- using the same criteria to determine their selection.
A summary is selected as the "chosen" one if it scores $\geq 4$ on the Likert scale or $\geq 80\%$ in percentage scores\footnotemark[4]. Conversely, a summary is considered "rejected" if its score is at least 1 point lower on the Likert scale or 20 percentage points lower than the chosen one.
To ensure a fair comparison, we standardize the number of pairs for each configuration to 92K, matching the number provided by C4. For clarity, we refer to the four summarizers as \emph{DPO-C1}, \emph{DPO-C2}, \emph{DPO-C3}, and \emph{DPO-C4}.

\footnotetext[4]{For multi-dimension cases (C3 and C4), we use a composite score, the average across all dimensions by default.}

\paragraph{Q2: Impact of Feedback Dimension.}

We examine how focusing on a specific dimension affects summary quality compared to considering all dimensions equally. This experiment provides insights into the concept of alignment tax \cite{noukhovitch2023language, guo2024controllable}, which refers to the trade-off where enhancing alignment with one objective (e.g., conciseness) may reduce performance in another (e.g., completeness). For each objective, we generate pairs of chosen and rejected summaries using the same criteria as in Q1. We then fine-tune Llama3-8b-instruct using DPO for preference learning, resulting in three compared summarizers: \emph{DPO-faith}, \emph{DPO-comp}, and \emph{DPO-cons}, each focusing on faithfulness, completeness, and conciseness, respectively.

\begin{table}[t]
\begin{center}
\footnotesize
\begin{tabular}{|L{1.2cm}| X{1.08cm} X{1.08cm} X{1.08cm} X{1.02cm}|} \toprule
~~~Level & C1 & C2 & C3 & C4 \\\midrule
\!Summary & 0.358 & 0.470 & 0.589 & 0.649 \\ 
\!System & 0.800 & 0.833 & 0.783 & 0.983 \\ \bottomrule
\end{tabular}
\end{center}
\vspace*{-0.4cm}
\caption{Quality of feedback using C1--C4 assessed based on their Spearman correlation with human composite scores in UniSumEval \cite{lee2023unisumeval}, evaluated at both the summary and system levels.} 
\label{table:metric-result}
\vspace*{-0.45cm}
\end{table}

\paragraph{Q3: Comparison of DPO over SFT.}

We compare the effectiveness of DPO with that of five SFT variants. The performance of SFT heavily depends on which summaries are used as reference for teacher-forcing \cite{han2024alignsumdatapyramidhierarchical}. We consider four different setups, namely \emph{SFT-human}: We fine-tune Llama3-8b-instruct using human summaries in our source datasets as reference; \emph{SFT-best}:  We first identify the best summary among those generated by 13 summary models in FeedSum, selecting the one with the highest composite score obtained from C4. Then, use it as the reference to supervise Llama3-8b-instruct; \emph{SFT-faith}, \emph{SFT-comp}, and \emph{SFT-cons}: Similar to SFT-best, but the best summary is selected based solely on their faithfulness, completeness, and conciseness scores, respectively.

\section{Analysis of LLM-generated Feedback}
\label{sec:feedback_analysis}

The configurations C1–C4 for LLM-based feedback generation show significant differences in feedback quality and their impact on constructing chosen-rejected summary pairs.

\subsection{Quality of LLM Feedback}

Table \ref{table:metric-result} presents the quality of LLM feedback by calculating summary- and system-levels agreement with human scores of UniSumEval (see Appendix \ref{appendix:metric-feedback} for the detailed metric). The feedback in C1 shows a significantly lower correlation with human scores compared to C2–C4, which is attributed to the use of a smaller LLM. The quality of feedback improves progressively as it becomes more multi-dimensional and fine-grained from C2 to C4.

\begin{table}[t]
\begin{center}
\footnotesize
\begin{tabular}{|X{0.9cm} |X{1.67cm} X{1.67cm} X{1.67cm}|} \toprule
Type & {Non-LLMs} & {Open. LLMs} & {Propr. LLMs} \\\midrule
C1 & 10.7 (60.7)  & 56.3 (29.6)  & 33.0 (9.7)  \\ 
C2 & ~~6.2 (74.4) & 59.4 (23.3)  & 34.4 (2.3) \\ 
C3 & ~~5.1 (81.2) & 60.9 (18.5)  & 34.0 (0.3)  \\ 
C4 & 19.3 (30.9)  & 55.1 (50.2)  & ~~25.6 (18.9)  \\ \bottomrule
\end{tabular}
\end{center}
\vspace*{-0.4cm}
\caption{Proportion (\%) of chosen (and rejected) summaries across three summarizer categories.} 
\label{table:chosen-rejected}
\vspace*{-0.5cm}
\end{table}

\begin{table*}[t]
\begin{center}
\footnotesize
\begin{tabular}{ |X{2.2cm} |X{1.2cm} |X{1.0cm} X{1.2cm} X{2.1cm} |X{1.1cm} X{1.1cm} X{1.1cm} |X{1.1cm}|} \toprule
Backbone & \!\!Config.\!\! & Quality & \!\!Dimension\!\! & Granularity &  Faith. & Comp. & Conc. & Avg. \\ \midrule
{Llama3-8b-inst.} & \multicolumn{4}{c|}{wo. RL} &  0.864 & 0.583 & 0.450 & 0.632 \\ \midrule
\multirow{4}{*}{Llama3-8b-inst.} & DPO-C1 & Low & Single & Coarse-grained & 0.836 & 0.594 & 0.453 & 0.628 \\
 & DPO-C2 & High & Single & Coarse-grained & 0.878 & 0.595 & 0.514 & 0.662 \\
& DPO-C3 & High & Multiple & Coarse-grained  & 0.889 & 0.581 & 0.516 & 0.662 \\
& DPO-C4 & High & Multiple & Fine-grained & \textbf{0.931} & \textbf{0.614} & \textbf{0.659} & \textbf{0.735} \\ \hline\midrule
Llama3-70b-inst.\!\! & \multicolumn{4}{c|}{wo. RL} & {0.931} & 0.596 & 0.487 & 0.671 \\ \midrule
Llama3-70b-inst.\!\! & DPO-C4 & High & Multiple & Fine-grained  & \textbf{0.950} & \textbf{0.632} & \textbf{0.754} & \textbf{0.779} \\\bottomrule
\end{tabular}
\end{center}
\vspace*{-0.4cm}
\caption{Automated evaluation results of seven summarizers on faithfulness, completeness, and conciseness. "w.o RL" refers to the models without preference optimization. "DPO-\{C1,C2,C3,C4\}" denote models incorporating preference optimization, utilizing feedback generated in C1--C4 of Table \ref{table:exp_configuration}. The best score are marked in bold.} 
\label{table:exp1}
\vspace*{-0.45cm}
\end{table*}

\subsection{Distribution of Summary Score}

Figure \ref{fig:feedback-distribution} illustrates the distribution of composite summary scores obtained from the four LLM-based evaluation strategies (C1--C4), as in Table \ref{table:exp_configuration}.

Firstly, \textbf{low-quality feedback from C1 tends to avoid assigning very low (score 1) or high (score 5) scores.} As shown in Figure \ref{fig:feedback-distribution}(a), most summaries receive scores between 3 and 4, regardless of the summarizer they were generated from. This results in the incorrect selection of low-quality summaries as the chosen ones.

Secondly, \textbf{feedback from the coarse-grained evaluation in C2–C3 introduces a significant bias, favoring LLM-generated summaries over non-LLM ones}, regardless of whether single- or multi-dimensional evaluation is used. Only a few LLM-generated summaries have scores below 4. This leads to the issue of indiscriminately selecting LLM summaries as chosen ones while non-LLM summaries as rejected ones. 

Lastly, \textbf{feedback from the fine-grained evaluation in C4 is robust to the summarizer category, showing a highly diverse score distribution across all summarizers.} This approach not only accurately captures the hierarchy among non-LLMs, open-source LLMs, and proprietary LLMs, but also demonstrates that high-quality summaries can be produced by older models. That is, summaries generated by LLMs can be rejected in favor of higher-quality ones produced by non-LLMs.


\footnotetext[5]{The overall trends in all experiments are consistent even with another LLM as the backbone. See the results with GPT-4o as the evaluator in Appendix \ref{sec:gpt-4o-eval}.}

\subsection{Chosen and Rejected Summary}

Table \ref{table:chosen-rejected} summarizes the proportion of summaries selected as chosen or rejected ones across the three summarizer categories. \textbf{A significant difference in proportion is observed depending on the feedback configuration.} 
Feedback from single-dimensional or coarse-grained automatic evaluations in C1–C3 predominantly classifies LLM-generated summaries as chosen in 89.3\%–94.9\% of cases, while non-LLM-generated summaries are rejected in 60.7\%–81.2\% of cases. However, feedback from the multi-dimensional and fine-grained evaluation in C4 reveals a considerably different trend, with 69.1\% of LLM-generated ones as rejected and 19.3\% of non-LLM-generated summaries being classified as chosen. This significant difference across configurations greatly affects their effectiveness in preference optimization.

\section{Main Experiment}
\label{sec:evaluation}
\vspace*{-0.05cm}

\paragraph{Test Set.}
The test set is constructed by randomly sampling 200 documents from the test split of \algname{}'s seven source datasets, totaling 1.4K. Both \emph{automated} and \emph{human} evaluations are conducted to evaluate summarizers on this set. For automated evaluation, we use Llama3-80b-instruct\footnotemark[5] as the backbone of FineSurE\,\cite{song2024finesure}. For human evaluation\footnotemark[6], we perform a fact verification task for faithfulness and a key-fact alignment task for completeness and conciseness, following the work\,\cite{lee2023unisumeval}.  Three annotators are assigned for each task, recruited through Amazon Mechanical Turk. Details of the automated and human evaluation can be found in Appendix \ref{appendix:human-eval-details}. 

\footnotetext[6]{Due to the high cost of fine-grained human evaluation, 420 document-summary pairs are randomly sampled, i.e., 10 (documents) $\times$ 7 (source data) $\times$ 6 (summarizers) in Table \ref{table:exp1}.}

\paragraph{Evaluation Metric.} We report the quality of text summary from the three key perspectives: \emph{faithfulness}, the proportion of faithful summary sentences; \emph{completeness}, the proportion of covered key-facts; and \emph{conciseness}, the proportion of summary sentences aligning with the key-facts. In automated evaluation, key-facts are automatically extracted from the reference summary of each source dataset, as suggested by \citet{song2024finesure}. In human evaluation, we use human-annotated key-facts from the UniSumEval dataset\,\cite{lee2023unisumeval} to follow the same manual evaluation pipeline in the paper. See Appendix \ref{appendix:metric-summary} for the equation to calculate three percentage scores. Appendix \ref{appendix:other-metric} presents the results using ROUGE and BERTScore for reference.

\subsection{Q1: Impact of Feedback Configuration}

We evaluate the quality of summaries generated by Llama3, trained with DPO incorporating feedback from four different configurations, and compare the results to the corresponding models without DPO. The prompts used to generate summaries are identical across all setups, as in Appendix \ref{sec:summary-prompt}. The results for other LLMs are in Appendix \ref{sec:exp-gemma}.

\begin{table}[t]
\begin{center}
\footnotesize
\begin{tabular}{|L{1.4cm} |X{1.02cm} X{1.02cm} X{1.02cm} |X{1.02cm}|} \toprule
Config. & Faith. & Comp. & Conc. & Avg. \\\midrule
\!Llama3-8b\!\! & 0.902 & 0.636 & 0.784 & 0.774 \\ 
\!Llama3-70b\!\! & 0.953 & 0.659 & 0.792 & 0.801 \\  \midrule 
\!DPO-C1 & 0.868 & 0.669 & 0.826 & 0.787 \\
\!DPO-C2 & 0.947 & 0.675 & 0.840 & 0.820 \\
\!DPO-C3 & 0.925 & 0.664 & 0.869 & 0.819 \\
\!DPO-C4 & \textbf{0.980} & \textbf{0.697} & \textbf{0.959} & \textbf{0.879} \\ \bottomrule
\end{tabular}
\end{center}
\vspace*{-0.424cm}
\caption{Human evaluation of six summarizers on three dimensions. Llama3-8b-instruct and Llama3-70b-instruct (without RL) are baselines, while C1–C4 are Llama3-8b-instruct after DPO with varying feedback.} 
\label{table:human-eval}
\vspace*{-0.0cm}
\end{table}

\begin{table}[t]
\begin{center}
\footnotesize
\begin{tabular}{|L{1.3cm} |X{1.55cm} X{1.55cm} X{1.55cm}| } \toprule
Strategy & Faith. & Comp. & Conc. \\\midrule
\!DPO-{faith}\!\!\! & \cellcolor{blue1} \!\!\textbf{0.942 {\scriptsize (+0.078)}}\!\!  & \!\!\cellcolor{blue3} 0.577 {\scriptsize (-0.006)}\!\! & \!\!\cellcolor{blue2}  0.686 {\scriptsize (0.236)}\!\! \\
\!DPO-comp\!\!\!\! & \cellcolor{blue4} \!\!0.846 {\scriptsize (-0.018)}\!\! & \cellcolor{blue1} \!\!\textbf{0.640 {\scriptsize (+0.057)}}\!\! & \cellcolor{blue4}  \!\!0.438 {\scriptsize (-0.012)}\!\! \\
\!DPO-cons\!\!\! & \cellcolor{blue3} \!\!0.877 {\scriptsize (+0.013)}\!\!  & \cellcolor{blue4}  \!\!0.493 {\scriptsize (-0.090)}\!\! & \cellcolor{blue1} \!\!\textbf{0.892 {\scriptsize (+0.442)}}\!\! \\ 
\!DPO-avg\!\!\! & \cellcolor{blue2} \!\!0.902 {\scriptsize (+0.038)}\!\! & \cellcolor{blue2}  \!\!0.608 {\scriptsize (+0.025)}\!\! & \cellcolor{blue3}  \!\!0.591 {\scriptsize (+0.141)}\!\!\\ \bottomrule
\end{tabular}
\end{center}
\vspace*{-0.424cm}
\caption{Automated evaluation results of summarizers trained with DPO on a specific dimension. DPO-avg is a model, averaging the LoRA weights of others. Cells are color-coded from dark to light based on descending score ranks for each dimension. The value in the parenthesis is gain or decline over the model wo. DPO.} 
\label{table:impact-dim}
\vspace*{-0.37cm}
\end{table}

\subsubsection{Results by Automated Evaluation}

Table \ref{table:exp_configuration} presents the summary quality of seven summarizers: two models without preference optimization and five models with DPO using LLM feedback generated in different configurations. 

\textbf{The low-quality feedback (C1) generated by Llama3-8b-instruct proves ineffective.} Compared to not using DPO, the faithfulness score rather drops by 0.028, resulting in the lowest composite score ("Avg.") across the three dimensions. 
The quality of feedback is crucial for preference learning using LLM feedback. \textbf{High-quality feedback (C2--C3) via coarse-grained evaluation improves the performance of summarizers in most cases.} However, there is no improvement in considering multi-dimensional aspects in automated evaluation as long as the granularity remains coarse. This is likely due to coarse-grained feedback lacking diversity in feedback pairs, indiscriminately selecting LLM-generated summaries as chosen and non-LLM summaries as rejected, as shown in Figures \ref{fig:feedback-distribution}(b) and (c). 
\textbf{To fully exploit LLM-generated feedback, fine-grained evaluation (C4) with diverse score distribution is crucial.} The improvement is significant, achieving a composite score of 0.735, which is 0.103 higher than Llama3-8b-instruct (w.o RL) and even surpasses the nearly 10x larger Llama3-70b-instruct (w.o RL).

Additionally, the improvement by fine-grained feedback (C4) even remains with Llama3-70b-instruct.
We name our best models SummLlama3-8b/70b and release them on Huggingface. Examples of summaries from different configurations are compared in Appendix \ref{sec:summ-example}.

\begin{table}[t]
\begin{center}
\footnotesize
\begin{tabular}{ |L{1.5cm} |X{0.7cm} X{0.75cm} X{0.7cm} X{0.7cm} |X{0.7cm}|} \toprule
Strategy &  \!\!Faith.\!\! & \!\!Comp\!\! & \!\!Conc\!\! & \!\!Abs.\!\! & \!\!Avg.\!\!\\ \midrule
Llama3-8b\!\!\! & 0.864\! & 0.583 & 0.450\! & 0.696\! & 0.648\!\\ 
DPO-C4\!\!\! & 0.931\! & 0.614 & 0.659\! & 0.691\! & 0.723\! \\ \midrule
SFT-human\!\!\!\!\! & 0.774\! & 0.496 & \textbf{0.666}\! & 0.516\! & 0.613\!\\
SFT-best\!\!\! & \underline{0.894}\! & \underline{0.551} & {0.572}\! & \textbf{0.588}\! & \textbf{0.651}\! \\
SFT-faith\!\!\! & \textbf{0.903}\! & 0.545 & 0.536\! & 0.559\! & {0.635}\!\\
SFT-comp\!\!\! & 0.871\! & \textbf{0.597} & 0.511\! & \underline{0.594}\! & \underline{0.643}\!\\
SFT-conc\!\!\! & 0.874\! & 0.483 &  \underline{0.632}\! & 0.488\! & {0.620}\! \\ \bottomrule
\end{tabular}
\end{center}
\vspace*{-0.424cm}
\caption{Comparison of DPO with SFT variants using automated evaluation, where they are fine-tuned from Llama3-8b-instruct. For SFT, the best score is marked in bold, while the second-best score is underlined.} 
\label{table:exp-q3}
\vspace*{-0.4cm}
\end{table}

\subsubsection{Results by Human Evaluation.}

Table \ref{table:human-eval} presents the results of the human evaluation across three dimensions. 
The overall performance dominance aligns with the automated evaluation results presented in Table \ref{table:exp1}. The DPO-C4, which is Llama3-8b-instruct fine-tuned with DPO using feedback from C4, significantly outperforms DPO-\{C1, C2, C3\}, and even surpasses the larger Llama3-70b-instruct. Therefore, the results from both automated and human evaluations confirm that \textbf{a smaller model can outperform its larger counterpart with appropriate training.}

\subsection{Q2: Impact of Feedback Dimension}

Table \ref{table:impact-dim} shows the summary quality of summarizers trained with DPO on a single feedback dimension from C4, along with the model obtained through post-hoc parameter merging\,\cite{jang2023personalized}.

Compared to the original Llama3-8b-instruct, DPO-\{faith, comp, cons\}, which rely on a single dimension, achieve the best scores in their target dimensions, but \textbf{they are likely to show limited improvements or even performance declines in other dimensions}. Specifically, focusing on completeness lowers faithfulness and conciseness, while prioritizing conciseness reduces completeness. 
Additionally, their parameter merging, DPO-avg, results in a balanced score across all dimensions, but training with chosen-rejected pairs based on the composite scores from the sixth row of Table \ref{table:exp1} (DPO-C4) achieves better results.

\subsection{Q3: Comparison of DPO over SFT}
\label{sec:sft-exp}
Table \ref{table:exp-q3} compares the summary quality of five SFT variants fine-tuned with reference summaries selected based on different policy, alongside Llama3-8b-instruct before and after DPO. 
Here, we introduce an additional dimension, "abstractiveness (Abs.)," which refers to the extent to which a summary generates novel sentences or phrases, leading to a more coherent summary\,\cite{zhang2022attention, song2023enhancing}.
The quality of generated summaries vary depending on the SFT policy.

Firstly, \textbf{DPO is a much superior approach to the SFT variants.} DPO-C4 significantly improves summary quality across multiple dimensions when compared to the vanilla Llama3-8b-instruct, while SFT-best falls short in delivering similar improvements, despite being fine-tuned with only the best-selected summaries during the training phase.
Secondly, \textbf{all SFT variants show a notable decline in abstractiveness}, incurring the \emph{copy bias} and leading to less coherent summaries due to sentence copying from the input document\,\cite{song2023enhancing}. This is because SFT allows only one reference summary per document, whereas DPO is superior by presenting multiple possible summaries through chosen-rejected pairs.
Thirdly, \textbf{focusing on a single dimension in SFT may improve that aspect but is likely to worsen others.} SFT-faith and SFT-cons improve their target dimensions compared to Llama3-8b-instruct but both sacrifice completeness. Thus, SFT-best, which equally considers all dimensions, achieves the highest average score ("Avg.") among the SFT variants.




\begin{table}[t]
\begin{center}
\footnotesize
\begin{tabular}{|X{0.85cm} |X{0.85cm} X{0.85cm} X{0.85cm} X{0.85cm} X{0.85cm}|} \toprule
Dim. & 0K & 13K\! & 23K\! & 46K & \!\!92K\!\!\\\midrule
Faith. & 0.864 & 0.914 & 0.937 & 0.930 & 0.931  \\
Comp.\! & 0.583 & 0.613 & 0.603 & 0.612 & 0.614  \\
Conc. & 0.450 & 0.594 & 0.611 & 0.648 & 0.659  \\\midrule
Avg. & 0.632 & 0.707 & 0.717 & 0.730 & 0.735 \\\bottomrule
\end{tabular}
\end{center}
\vspace*{-0.4cm}
\caption{Automated evaluation results as the number of selected-rejected summary pairs of LLM-generated feedback for DPO increases from 0 to 92K.} 
\label{table:feedback-size}
\vspace*{-0.45cm}
\end{table}

\subsection{Additional Experiment}
\label{sec:add-exp}

\paragraph{Feedback Size.} We explore the impact of varying the size of high-quality, multi-dimensional, fine-grained feedback generated by LLMs in C4. The percentage scores, based on the number of selected-rejected pairs used for DPO, are presented in Table \ref{table:feedback-size}.
Overall, {the scores gradually improve as the amount of feedback pairs increases.} Notably, there is a significant improvement with as few as 13K summary feedback pairs, but the increase nearly plateaus after 46K feedback pairs. Thus, high-quality, multi-dimensional, fine-grained feedback is essential for using LLM feedback in text summarization, {with around 50K feedback pairs being a reasonable fit for preference learning}.

\begin{table}[t]
\begin{center}
\footnotesize
\begin{tabular}{|X{1.4cm} |X{0.4cm} |X{0.8cm} X{0.8cm} X{0.85cm} |X{0.8cm}|} \toprule
Source & \!\!\!Size\!\!\! & Faith. & \!Comp.\! & Conc. & Avg. \\\midrule
\!\!UniSumEval\!\!\! & 1K & 0.874 & 0.618 & 0.506 & 0.666 \\
\!\!SynFacEdit\!\!\! & 5K & 0.789 & 0.520 & 0.563 & 0.624 \\ \midrule
\multirow{2}{*}{\makecell{\!\!FeedSum\!\! \\(Ours)}} & 5K & 0.913 & 0.606 & 0.587 & 0.702 \\ 
&  \!\!92K\!\! & 0.931 & 0.914 & 0.659 & 0.735 \\ \bottomrule
\end{tabular}
\end{center}
\vspace*{-0.4cm}
\caption{Effectiveness of feedback in FeedSum over UniSumEval (human feedback) and SynFacEdit (synthetic feedback). "Size" refers to the available number of pairs consisting of chosen and rejected summaries.} 
\label{table:eval-feedback}
\vspace*{-0.3cm}
\end{table}

\paragraph{Human and Synthetic Feedback.} 
In Table \ref{table:eval-feedback}, we compare the effectiveness of using FeedSum's feedback from C4 with (1) human feedback on real summaries in UniSumEval \cite{lee2023unisumeval}, and (2) synthetic feedback on synthesized summaries in SynFacEdit \cite{mishra2024synfac}. We train Llama3-8b-instruct using DPO, but with different feedback from the three datasets.
%
Despite the limited size, DPO with human feedback in UniSumEval improves summary quality across all dimensions, raising the average score from 0.632 to 0.666 compared to Llama3-8b-instruct without DPO. 
Although the feedback in FeedSum is obtained through automated evaluation, it shows greater improvements than using 1K human feedback. Enhancements are more pronounced when increasing the size of feedback. 
On the other hand, the synthetic feedback in SynFacEdit decreases the quality of generated summaries after DPO, likely due to its limited size and focus on clinical summarization.

\paragraph{Optimization with PPO and KTO.} 

\begin{table}[t]
\begin{center}
\footnotesize
\vspace*{0.224cm}
\begin{tabular}{|L{1.4cm} |X{1.02cm} X{1.02cm} X{1.02cm} |X{1.02cm}|} \toprule
Config. & Faith. & Comp. & Conc. & Avg. \\\midrule
\!wo. RL\!\! & 0.864 & 0.583 & 0.450 & 0.632 \\ \midrule 
\!DPO-C4 & 0.931 & 0.641 & 0.659 & 0.735 \\
\!PPO-C4 & 0.842 & 0.558 & 0.426 & 0.619 \\
\!KTO-C4 & 0.809 & 0.593 & 0.788 & 0.730 \\ \bottomrule
\end{tabular}
\end{center}
\vspace*{-0.404cm}
\caption{Comparison of PPO and KTO with DPO using C4 Feedback for preference optimization.} 
\label{table:optimization-methods}
\vspace*{-0.4cm}
\end{table}

We obtain results using KTO \cite{ethayarajh2024kto} under the exact same training setup as DPO, while using PPO \cite{schulman2017proximal} under different setup due to its requirement of a reward model (see Appendix \ref{sec:train-configuration}).
Table \ref{table:optimization-methods} shows the summary quality improvements achieved by DPO, PPO, and KTO with Llama3-8B-inst using the C4 feedback. 

Firstly, the summary quality of using PPO is significantly worse than using DPO, likely due to the reward model's difficulty in accurately assessing summaries during PPO training, as mapping multi-dimensional scores to summaries using the Llama3-8b-instruct-based reward model is challenging. In contrast, DPO benefits from directly using high-quality feedback from Llama3-70B-instruct without training a reward model. 

Secondly, the results demonstrate that while KTO is comparable to DPO in terms of the average score, it tends to compromise faithfulness significantly to achieve a substantial improvement in conciseness. We believe this reflects the tendency of absolute-criteria methods (as in KTO) to focus on the most vulnerable evaluation dimension, such as conciseness in this case, rather than balancing trade-offs like pairwise comparison methods.




\vspace*{-0.1cm}
\section{Conclusion}
\label{sec:conclusion}
\vspace*{-0.1cm}


This work presents a framework for improving text summarization using LLM-generated feedback. We demonstrate that this approach is the most effective when the feedback is high-quality, multi-dimensional, and assessed at a fine-grained level. Our experiments show that DPO significantly outperforms SFT variants in utilizing such feedback. Additionally, we provide insights into the alignment trade-offs in summarization, the impact of feedback size, and the advantages of our LLM-generated feedback over existing human and synthetic alternatives. As part of our contribution, we open-sourced both the FeedSum dataset and the SummLlama model on Hugging Face. 

\section*{Limitations}\label{sec:lim}

DPO is a widely used approach for preference optimization; however, it has limitations in handling multi-dimensional feedback. A typical method involves computing a composite score by averaging the scores across all dimensions with equal weights, which may not be the optimal solution for multi-dimensional preference learning. Although we include a baseline of post-hoc parameter merging \cite{jang2023personalized}, recent work suggests there are better performance alternatives, such as Controllable DPO \cite{guo2024controllable} and Sequential Alignment \cite{lou2024spo}. We will explore the extent of performance improvement achieved by these solutions in future work.

We conducted both human and automated evaluations. However, the majority of the evaluations were automated due to the high cost associated with fine-grained, multi-dimensional manual assessments. Nevertheless, we believe that the automated evaluations provide convincing evidence, as they have demonstrated performance comparable to human evaluations \,\cite{song2024finesure, tang2024minicheck, liu2023g}.

\section*{Ethics Statement}\label{sec:ethics}

Our work primarily focuses on leveraging LLM-generated feedback on diverse text summaries, which does not pose any ethical concerns during the model training phase. For human evaluation, we followed a well-defined evaluation protocol in the literature, preventing possible ethical issues in the annotation process. Annotators were paid 50\% more than the average U.S. minimum wage and received bonuses for maintaining consistent, high-quality performance.

\section*{Scientific Artifacts}\label{sec:sci}

The summaries used to collect LLM feedback were generated by 13 different language models. For open-source models, we used publicly available checkpoints from Huggingface, while for proprietary models, we utilized paid API services provided by OpenAI and AWS Bedrock. See Table \ref{tab:summarizer-source} for details in Appendix.

\section*{Acknowledgements}\label{sec:ack}


This work was supported by Institute of Information \& communications Technology Planning \& Evaluation (IITP) grant funded by the Korea goverment (MSIT) (No. RS-2024-00445087, Enhancing AI Model Reliability Through Domain-Specific Automated Value Alignment Assessment). Additionally, this work was partly supported by the National Research Foundation of Korea (NRF) grant funded by the Korea government (MSIT) (No. RS-2024-00334343) and by Artificial Intelligence industrial convergence cluster development project funded by the Ministry of Science and ICT (MSIT, Korea) \& Gwangju Metropolitan City (No. BA00001698).




\appendix
\clearpage

\begin{table}[t!]
\footnotesize
\begin{center}
\setlength{\tabcolsep}{4.5pt}
\renewcommand{\arraystretch}{1.1}
\begin{tabular}{|ll|}
\toprule
\multicolumn{1}{|l}{Model   Name}        & Checkpoints   \\ \midrule
Bart-large   
& facebook/bart-large-cnn  \\
Pegasus-large  & google/pegasus\-cnn\_dailymail  \\
Flan-t5-large & spacemanidol/flan-t5-large-cnndm     \\
Phi-2                          
& microsoft/phi-2                      \\
Mistral-7b-inst           
& mistralai/Mistral-7B-Instruct-v0.2   \\
Mixtral-8x7b-inst           
& mistralai/Mixtral-8x7B-Instruct-v0.1 \\
Llama2-13b-chat                
& meta-llama/Llama-2-13b-chat-hf   \\
Mistral-nemo               
& mistralai/Mistral-Nemo-Instruct-2407  \\
Gemma2-27b-inst              
& google/gemma-2-27b-it   \\
Llama3-70b               
& meta-llama/Meta-Llama-3-70B-Instruct    \\
Claude-instant                      
& claude-instant (AWS Bedrock)            \\ 
GPT-3.5\textsubscript{turbo}             
& gpt-3.5-turbo-0125 (OpenAI)            \\
GPT-4\textsubscript{turbo}                          
& gpt-4-0125-preview (OpenAI)           \\
\bottomrule
\end{tabular}
\vspace*{-0.25cm}
\caption{Checkpoints of the 13 summarizers. For open-source models, we use publicly available checkpoints from Huggingface, while for proprietary models, we utilize paid API services by OpenAI and AWS Bedrock.}
\label{tab:summarizer-source}
\end{center}
\vspace*{-0.3cm}
\end{table}

\begin{table}[t]
\begin{center}
\footnotesize
\begin{tabular}{|L{7.2cm}|} \toprule
You will receive an article along with a summary of that article. 

\vspace*{0.2cm}
Please evaluate the quality of summary on a Likert-scale score from 1 (bad) to 5 (perfect).

\vspace*{0.2cm}
Provide your answer in JSON foramt. The answer should be a dictionary whose key is "score": 

\{"score": "your score"\}

\vspace*{0.2cm}
Source Text: 

{\color{blue}\{source text\}}

\vspace*{0.2cm}
Summary: 

{\color{blue}\{summary\}}

\vspace*{0.2cm}
JSON Output:\\ \bottomrule
\end{tabular}
\end{center}
\vspace*{-0.4cm}
\caption{Prompt to generate low-quality, single-dimensional, fine-grained feedback using C1.} 
\label{table:prompt-c1}
\vspace*{-0.4cm}
\end{table}

\begin{table}[t]
\begin{center}
\footnotesize
\begin{tabular}{|L{7.2cm}|} \toprule
You will be provided with a transcript. Your task is to decompose the summary into a set of "key facts". 

A "key fact" is a single fact written as briefly and clearly as possible, encompassing at most 2-3 entities.

\vspace*{0.2cm}
Here are nine examples of key facts to illustrate the desired level of granularity:

* Kevin Carr set off on his journey from Haytor.

* Kevin Carr set off on his journey from Dartmoor.

* Kevin Carr set off on his journey in July 2013.

* Kevin Carr is less than 24 hours away from completing his trip.

* Kevin Carr ran around the world unsupported.

* Kevin Carr ran with his tent.

* Kevin Carr is set to break the previous record.

* Kevin Carr is set to break the record by 24 hours.

* The previous record was held by an Australian.

\vspace*{0.2cm}
Instruction:

First, read the summary carefully.
Second, decompose the summary into (at most 16) key facts.

\vspace*{0.2cm}
Provide your answer in JSON format. The answer should be a dictionary with the key "key facts" containing the key facts as a list:

\{"key facts": ["first key fact", "second key facts", "third key facts"]\}

\vspace*{0.2cm}
Summary:
{\color{blue}\{summary\}}

JSON Output:\\\bottomrule
\end{tabular}
\end{center}
\vspace*{-0.4cm}
\caption{Prompt to extract the list of key-facts from the reference (human) summary of original datasets.} 
\label{table:key-fact-extraction}
\vspace*{-0.5cm}
\end{table}

\section{Data Creation Details}
\label{sec:feedback-generation}

\begin{table*}
\begin{center}
\scriptsize
\begin{tabular}{|X{1.6cm}|X{1.4cm}|X{1.2cm}|X{1.3cm}|X{1.4cm}|X{2.1cm}|X{2.0cm}|X{1.4cm}|}
\toprule
Dataset &
Type &
\makecell{\!\!\!\!Document\!\!\!\! \\ Length} &
Domain &
\makecell{\# of \\Document} &
\makecell{Document \\ Word count \\ (Min -- Max)} &
\makecell{Summary \\ Word count \\ (Min -- Max)} &
\makecell{\!\!\!\!\!\!\!\!Key-fact Count\!\!\!\!\!\!\!\! \\ (Min -- Max)} \\ \midrule
CNNDM       & \multirow{4}{*}{\makecell{\\ \\ \!\!\!\!\!Non-Dialogue\!\!\!\!\!}} & \multirow{2}{*}{\makecell{Short}} & News      & 21194 & \makecell{\!\!\!\!675.8 (46--1919)\!\!\!\!} & \makecell{\!\!\!\!48.8 (10--162)\!\!\!\!} & \makecell{\!\!6.3 (1--16)\!\!} \\ 
\cmidrule(lr){1-1} \cmidrule(lr){4-8}
WikiHow     &                                        &                        & Lifestyle   &  21980 & \makecell{72.9 (10--680)}  & \makecell{6.56 (1--51)}   & \makecell{1.4 (1--16)}   \\ 
\cmidrule(lr){1-1} \cmidrule(lr){3-8}
\!\!GovReport\!\!   &                                        & \multirow{2}{*}{\makecell{\\ Long}}  & Report     &  8066 & \makecell{\!\!3573.0 (141--5873)\!\!} & \makecell{\!\!\!\!\!\!439.4 (29--1002)\!\!\!\!\!\!} & \makecell{14.7 (4--21)} \\ 
\cmidrule(lr){1-1} \cmidrule(lr){4-8}
PubMed      &                                        &                        & Medical     &  17843 & \makecell{2491.0 (10--6384)} & \makecell{\!\!\!\!210.6 (49--402)\!\!\!\!} & \makecell{12.0 (3--26)} \\ 
\midrule
\!\!DialogSum\!\!  & \multirow{3}{*}{\makecell{\\ \\ Dialogue}}              & \multirow{1}{*}{\makecell{ Short}} & Daily Life  & 21957 & \makecell{122.4 (33--727)} & \makecell{22.6 (5--101)} & \makecell{3.7 (1--16)} \\ 
\cmidrule(lr){1-1} \cmidrule(lr){3-8}
\!\!MediaSum\!\!    &                                        & \multirow{2}{*}{\makecell{\\ Long}}  & Interview   &  18927 & \makecell{1373.7 (80--5111)} & \makecell{14.3 (5--97)} & \makecell{2.6 (1--14)} \\ 
\cmidrule(lr){1-1} \cmidrule(lr){4-8}
\!\!\!\!\!MeetingBank\!\!\!\!\! &                                        &                        & Meeting  & 15421  & \makecell{1283.1 (96--5803)} & \makecell{56.4 (14-184)}  & \makecell{6.8 (1--20)} \\ \bottomrule
\end{tabular}
\vspace{-0.2cm}
\caption{Statistics of the \algname{} training set, detailing the average word count of input documents, reference summaries, and key-facts, with respective min-max ranges in parentheses (reference summaries refer to the human-written summaries in the original datasets). Documents with over 1K words are considered "long".}
\label{table:source_datasets}
\end{center}
\vspace{-0.2cm}
\end{table*}

\subsection{Feedback Generation}

We generate LLM-based feedback across four different setups, as summarized in Table \ref{table:exp_configuration}:
\begin{itemize}
\item \textbf{C1:} This setup is designed to acquire low-quality, coarse-grained, single-dimensional feedback. We perform automated evaluation using the prompt in Table \ref{table:prompt-c1} with Llama3-8-instruct, a lower-performing model compared to its larger counterpart, Llama3-70b-instruct. The feedback obtained is a Likert-scale overall score for the summary.
\item \textbf{C2:} The prompt for this setup is identical to that of C1, but we use a nearly 10$\times$ larger LLM, Llama3-70b-instruct, to generate high-quality, single-dimensional, and coarse-grained feedback. 
\item \textbf{C3:} We use G-Eval \cite{liu2023g} with simple modification to tune it for our three key dimensions, namely faithfulness, completeness, and conciseness. We perform automated evaluation using the three prompts in Table \ref{table:modified-geval}. The feedback obtained is three Likert-scale scores for the three dimensions.
\item \textbf{C4:} We use FineSurE \cite{song2024finesure} to acquire high-quality, multi-dimensional, and fine-grained feedback at the sentence level for faithfulness; and at the key-fact level for completeness and conciseness. It performs a fact-checking task for the former and a key-fact alignment task for the latter using LLMs. The prompts for the two tasks are presented in Table \ref{table:finesure-prompt}. The feedback obtained is three percentage (\%) scores for the three dimensions.
\end{itemize}

\subsection{Key-fact Extraction}

The feedback from C4 requires fine-grained evaluation using key-facts to assess the completeness and conciseness scores.
The key-facts are automatically extracted from the reference (human) summary of each source dataset, as suggested by \citet{song2024finesure}. 
Thus, we obtain the list of key-facts for 15.4K documents in FeedSum: 14K for training set and the remaining 1.4K for testing set. The prompt used for automated key-fact extraction is detailed in Table \ref{table:key-fact-extraction}.

\subsection{Dataset Statistic}

We present a comprehensive statistical analysis of the FeedSum datasets, which consist of 125,388 <document, summary, feedback> triplets for each configuration outlined in Table \ref{table:exp_configuration}. Detailed statistics of FeedSum are provided in Table \ref{table:source_datasets}.


\section{Training Detail}
\label{sec:train-configuration}

\subsection{Training Configuration}

For preference learning, we investigate two possible solutions of supervised fine-tuning (SFT) and direct preference optimization (DPO). The details of each configuration are detailed below:

\paragraph{Supervised Fine-tuning (SFT).} We fine-tune Llama3-8b-instruct using QLoRA\,\cite{dettmers2024qlora} and DeepSpeed (Stage-2)\,\cite{rasley2020deepspeed} on four NVIDIA H100 GPUs. The model is trained for 3,000 steps with AdamW as the optimizer, using a batch size of 32, an initial learning rate of 1e-4, and a weight decay of 0.05. Regardless of how to select the reference summary, we apply the same configuration for all SFT strategies, namely SFT-\{human, best, faith, comp, conc\} in Table \ref{table:exp-q3}. 
The input (user prompt) and output (assistant prompt) for Llama3-8b-instruct are configured similarly to the example for DPO in Table \ref{table:dpo-prompt-example}. The difference is that SFT only passes the input along with a single output, selected based on a predefined criterion, e.g., the summary with the highest composite score.

\paragraph{Direct Preference Optimization (DPO).} We train Llama3-8b/70b-instruct using DPO \cite{rafailov2024direct}. Since the model has completed the instruction-tuning process, we proceed directly to optimize it using DPO. Like SFT, we apply QLoRA and DeepSpeed (Stage-2) to train the model on four NVIDIA H100 GPUs. The model is trained for 6,000 steps with AdamW as the optimizer, using a batch size of 32, an initial learning rate of 5e-5, and a weight decay of 0.05. We apply the same setup for all configurations, namely C1 -- C4 in Table \ref{table:exp_configuration}. The input (user prompt) and output (assistant prompt) for Llama3-8b-instruct are configured in the example of Table \ref{table:dpo-prompt-example}.

For ablation studies for feedback type and size in Section \ref{sec:add-exp}, we adjust the number of steps due to the different number of human or LLM-generated feedback. We reduce the number of steps to 4,000 when the number of feedback pairs exceeds 40,000; otherwise, we reduce it to 3,000. Other configurations remain the same.

\paragraph{Proximal Policy Optimization (PPO).} 
Firstly, reward model is trained Using the 92K pairwise feedback dataset in the DPO-C4 setting. We trained the reward model based on the Llama3-8b-instruct model for 15,000 steps. However, we observe that the accuracy of the reward scores are suboptimal, likely due to the limitations of the small base model (Llama3-8b-instruct). Specifically, the correlation between the reward scores from this model and evaluations using a larger Llama3-70b-instruct model is only 0.781 on our 1,400 test set in FeedSum.
Secondly, regarding PPO Results, we conduct experiments with PPO using the trained reward model for 30,000 steps (five times more than our DPO setups). 

\subsection{Input and Output Format}
Tables \ref{table:dpo-prompt-example} presents an example of the input and its corresponding chosen and rejected outputs to train Llama3-8b-instruct using DPO. We follow the same prompt style of Llama3 for instruction tuning. In this example, the chosen summary was generated by GPT-4-turbo, achieving scores of 100\% for faithfulness, 60\% for completeness, and 100\% for conciseness. On the other hand, the rejected summary was generated by Mistral-7b-instruct, achieving scores of 25\% for faithfulness, 60\% for completeness, and 50\% for conciseness. 

The auto-evaluation results of the example by FineSurE\,\cite{song2024finesure} are provided in Table \ref{table:ex-finesure}. The final percentage (\%) scores can be computed by calculating the proportion of factually correct sentences for faithfulness, that of included given key-facts for completeness, and that of summary sentences related to the key-facts. The detailed equation is provided in Appendix \ref{appendix:metric-summary}.

For SFT, the input and its corresponding response are almost similar to those of DPO. But, there is no distinction between chosen and rejected summaries. We select reference summaries using five different criteria: SFT-human, SFT-best, SFT-faith, SFT-comp, and SFT-cons, as detailed in Section \ref{sec:preference-learning}. The selected reference summaries are provided to train Llama3 using teacher-forcing.

\section{Automated and Human Evaluation}
\label{sec:detail-evaluation}

\begin{table*}[t]
\begin{center}
\footnotesize
\begin{tabular}{ |X{1.4cm} |X{2.5cm} |X{1.43cm} X{1.43cm} X{1.43cm} X{1.43cm} X{1.43cm} X{1.43cm}|} \toprule
Method & Backbone & \!ROUGE-1\! & \!ROUGE-2\! & \!\!ROUGE-L\!\! & BERT-F1 & BERT-P & BERT-R\\ \midrule
wo. RL & Llame3-8b-inst.  & 0.453 & 0.172 & 0.231 & 0.854 & 0.841 & 0.867 \\
wo. RL & Llame3-70b-inst. & 0.450 & 0.183 & 0.213 & 0.854 & 0.843 & 0.867 \\ \midrule
DPO-C1 & Llame3-8b-inst.  & 0.435 & 0.182 & 0.240 & 0.853 & 0.840 & 0.866 \\
DPO-C2  & Llame3-8b-inst. & \textbf{0.480} & 0.187 & \textbf{0.257} & 0.856 & 0.846 & 0.867 \\
DPO-C3  & Llame3-8b-inst. & 0.419 & 0.168 & 0.231 & 0.856 & \textbf{0.847} & 0.866 \\
DPO-C4  & Llame3-8b-inst. & 0.474 & \textbf{0.188} & 0.234 & \textbf{0.857} & 0.846 & \textbf{0.869} \\ \bottomrule
\end{tabular}
\end{center}
\vspace*{-0.4cm}
\caption{Results using two conventional automated metrics on six summarizers with and without DPO: ROUGE-\{1, 2, L\} and BERTScore-\{F1, Precision, Recall\}. The best scores are marked in bold.} 
\label{table:conventional-metric}
\vspace*{-0.3cm}
\end{table*}

\subsection{Metric for Summary Quality}

We utilize three dimensions of metrics, namely faithfulness, completeness, and conciseness, along with one that estimates the abstractiveness of the summary, in line with recent literature \cite{song2024finesure, lee2023unisumeval}.

\label{appendix:metric-summary}

\paragraph{Faithfulness Score.}

Faithfulness score is formulated by aggregating sentence-level fact check results. Let \( S = \{s_{1}, \ldots, s_{N}\} \) is the summary passage which consists of \(N\) sentences, where \( s_{i} \) denotes the \(i\)-th sentence in the summary passage. Let \( S_{\text{fact}} \subseteq S \) represent the subset of sentences verified as "factually correct." The faithfulness percentage score of \( S \), with respect to the document \( D \), is defined as:
\begin{equation}
\mathrm{Faithful}(D, S) = {|S_{\text{fact}}|}/{|S|}.
\end{equation}

This metric measures the proportion of factually correct sentences in the summary relative to the total number of sentences in the summary.

\paragraph{Completeness and Conciseness Score.}
Let \( K = \{ k_1, \ldots, k_M \} \) be the collection of key-facts, where \( M \) indicates the total number of these facts. Utilizing the results from the alignment of key-facts, we can establish a bipartite graph \( M = (K, S, E) \), with set of edges \( E = \{ (k, s) : k \to s \, | \, k \in K \land s \in S \} \). Here, the notation \( k \to s \) signifies that the key-fact \( k \) is identified as being included in the summary sentence \( s \). The completeness and conciseness scores for summary \( S \) are computed as percentage scores, defined as follows:
\begin{equation}
\mathrm{Complete}(K, S) = {| \{ k \, | \, (k, s) \in E \} |}/{|K|},
\end{equation}
\begin{equation}
\mathrm{Concise}(K, S) = {| \{ s \, | \, (k, s) \in E \} |}/{|S|}.
\end{equation}

In this context, the operator \( | \cdot | \) denotes the cardinality of a set. Completeness score indicates how well the key-facts are incorporated into the summary. Furthermore, the conciseness score evaluates how effectively the summary condenses and includes the key-facts.

\paragraph{Composite Score.}
To determine the chosen and rejected summaries in cases of multi-dimensional feedback, we use the average of the three percentage scores -- faithfulness, completeness, and conciseness -- to calculate a composite score.

\paragraph{Abstractiveness Score.}
In Section \ref{sec:sft-exp}, we additionally report the abstractiveness score of the summary, which refers to the extent to which a summary generates novel sentences or phrases, leading to a more coherent summary.
The abstractiveness score is measured by calculating the ratio of novel n-grams present in the summary that does not appear in the original input text \cite{liu2019text, song2023enhancing}. Let \( n\text{-gram}_{\text{shared}} \) represent the set of n-grams that are shared between the summary and the document, while \( n\text{-gram}_{\text{summary}} \) denotes the total set of n-grams included in the summary. Then, the ratio of novel n-grams \( N_n \) is defined as:
\begin{equation}
N_n = 1 - {|n\text{-gram}_{\text{shared}}|}/{|n\text{-gram}_{\text{summary}}|}.
\end{equation}

The final abstractiveness score for a summary \( S \) is computed as the average of the novel 1/3/5-gram ratios, as follows:
\begin{equation}
\mathrm{Abstractive}(D, S) = {(N_1 + N_3 + N_5)}/{3}.
\end{equation}

\subsection{Metric for Feedback Quality}
\label{appendix:metric-feedback}

In table \ref{table:metric-result}, we use the same settings as in recent studies (\citealt{liu2023improving}, \citealt{song2024finesure}) to evaluate the summary feedback quality and align it with human judgment.
Specifically, there are two levels for evaluating the alignment (correlation) of generated summary feedback with human feedback. The greater the alignment, the higher the quality of the generated feedback.

\paragraph{Summary-level Correlation.}

We can check the alignment between the generated and human feedback at the summary level.
To calculate the {summary-level correlation}, we define \( F_{actual} \) and \( F_{pred} \) as the percentage scores of the ground truth and the predicted summaries, respectively. Let \( D = \{D_1, \ldots, D_k\} \) represent the set of input documents, and \( S = \{S_1, \ldots, S_k\} \) represent the corresponding summaries for these documents. The summary-level correlation is computed as:
\begin{equation}
\begin{gathered}
\mathrm{Spearman}([F_{actual}(D_{1}, S_{1}), \ldots, F_{actual}(D_{k},S_{k})], \\ [F_{pred}(D_{1}, S_{1}), \ldots, F_{pred}(D_{k}, S_{k})]).
\end{gathered}
\end{equation}

We employ the Spearman correlation as our correlation measure. Ultimately, the summary-level correlation reflects the alignment between human-assessed feedback and LLM-generated feedback for the identical document.

\paragraph{System-level Correlation.}
The system-level evaluation assesses the alignment of performance rankings across summarization systems (summarizers) as determined by both our LLM feedback scores and human feedback scores.
To calculate the {system-level rank correlation}, we consider \( \mathbf{F}_m = \{F_m(D_1,S_1), \ldots, F_m(D_M,S_M)\} \) as the set of percentage scores derived from \( M \) document-summary pairs generated by the summarization model \( m \). Next, we construct a list of the average percentage scores for all \(T\) summarization models, denoted as \( \left[\bar{\mathbf{F}}_{m_1}, \ldots, \bar{\mathbf{F}}_{m_T} \right] \) where, \( \bar{\mathbf{F}}_m = \frac{1}{M}\sum_{i=1}^{M}{F}_{m}(D_{i},S_{i})\). Applying the rank function to this list, we derive the ranking list \(\mathrm{rank}\left([\bar{\mathbf{F}}_{m_1}, \ldots, \bar{\mathbf{F}}_{m_T}]\right) = \left[\mathrm{rank}_{m_1},\dots,\mathrm{rank}_{m_T}\right]\), where \( \mathrm{rank}_{m} \) represents the rank of model \( m \):
\begin{equation}
\mathrm{rank}(\bar{\mathbf{F}}_{m_i}) = \sum_{j=1}^{T} \mathbf{1}(\bar{\mathbf{F}}_{m_j} \leq \bar{\mathbf{F}}_{m_i}). 
\end{equation}

We derive the rank list \( \left[\mathrm{rank}_{m_1},\dots, \mathrm{rank}_{m_T} \right] \) based on LLM feedback scores, as well as the rank list \( \left[\mathrm{rank}_{m_1}^*,\dots, \mathrm{rank}_{m_T}^* \right] \) based on human feedback scores. Then, the system-level correlation is computed as:
\begin{equation}
\begin{gathered}
\mathrm{Spearman}(\left[\mathrm{rank}_{m_1},\dots, \mathrm{rank}_{m_T}\right], \\
\left[\mathrm{rank}_{m_1}^*,\dots, \mathrm{rank}_{m_T}^*\right]).
\end{gathered}
\end{equation}

The system-level rank correlation evaluates the degree of agreement between the rankings generated from LLM feedback scores and human-provided feedback scores across different summarization systems.

\subsection{Human Evaluation Details}
\label{appendix:human-eval-details}

\paragraph{Fine-Grained Annotation Tasks}
We conduct two human annotation tasks: (1) fact verification and (2) key-fact alignment. The format of the two annotation tasks draw on the annotation protocol suggested by \citet{lee2023unisumeval}.  In fact verification, annotators assign a binary label (Yes/No) to indicate whether a sentence contains factual errors.
For key-fact alignment, annotators evaluate whether summary sentences contain key-fact of their source text. We use the human-verified key-facts from the existing dataset created by \citet{lee2023unisumeval}.

For the two annotation tasks, we compute percentage scores for three summary-level evaluation dimensions: (1) faithfulness, the proportion of factually accurate sentences; (2) completeness, the percentage of key-facts covered by the summary; and (3) conciseness, the proportion of sentences relevant to the key-facts. The detailed formulation can be found in Appendix \ref{appendix:metric-summary}.

\paragraph{Annotator Qualifications and Costs}
We used Amazon Mechanical Turk (MTurk) annotators with an approval rating above 95\% and at least 1,000 accepted HITs. A detailed qualification test of English comprehension questions, simulating the actual annotation tasks, was required. We only recruited annotators who received the perfect score on the test and resided in AU, CA, NZ, GB, or the US. The total cost of human annotation exceeded \$750 for 420 input text-summary pairs, with payments above the U.S. minimum wage.

\begin{table*}[t]
\begin{center}
\footnotesize
\begin{tabular}{ |X{2.2cm} |X{1.2cm} |X{1.0cm} X{1.2cm} X{2.1cm} |X{1.1cm} X{1.1cm} X{1.1cm} |X{1.1cm}|} \toprule
Backbone & \!\!Config.\!\! & Quality & \!\!Dimension\!\! & Granularity &  Faith. & Comp. & Conc. & Avg. \\ \midrule
{Gemma-2b-inst.} & \multicolumn{4}{c|}{wo. RL} &  0.558 & 0.361 & 0.422 & 0.447 \\ \midrule
\multirow{4}{*}{Gemma-2b-inst.} 
& DPO-C1 & Low & Single & Coarse-grained & 0.507 & 0.373 & 0.463 & 0.448 \\
& DPO-C2 & High & Single & Coarse-grained & 0.556 & 0.383 & 0.498 & 0.479 \\
& DPO-C3 & High & Multiple & Coarse-grained  & 0.588 & 0.384 & 0.481 & 0.484 \\
& DPO-C4 & High & Multiple & Fine-grained & \textbf{0.613} & \textbf{0.396} & \textbf{0.533} & \textbf{0.514} \\ \bottomrule
\end{tabular}
\end{center}
\vspace*{-0.4cm}
\caption{Automated evaluation results of five summarizers on faithfulness, completeness, and conciseness. The model was initialized from the \textbf{Gemma-2b-instruct backbone} and trained using DPO. "w.o RL" refers to the models without preference optimization. "DPO-\{C1,C2,C3,C4\}" denote models incorporating preference optimization, utilizing feedback generated in C1--C4 of Table \ref{table:exp_configuration}. The best scores are marked in bold.} 
\label{table:exp-dpo-gemma}
\vspace*{-0.05cm}
\end{table*}

\section{Results with Conventional Metric}
\label{appendix:other-metric}

We conduct automated evaluation using two popular conventional metrics, namely ROUGE \cite{lin2004rouge} and BERTScore \cite{zhang2019bertscore}. Although it has been recently recognized that these metrics do not align well with human evaluations of text summaries \cite{pagnoni2021understanding, song2024finesure}, these scores can still serve as auxiliary metrics to assess word overlap (using ROUGE) and semantic relevance (using BERTScore) with the given reference summaries.

Table \ref{table:conventional-metric} presents the ROUGE and BERT scores on the test set of FeedSum. We use the human-written summaries from the six source datasets as reference summaries to compute the respective scores. All the summarizers demonstrate consistently high ROUGE and BERT scores, both with and without DPO. While there is no significant difference between them, DPO-C4 achieves the highest scores in three out of six score categories.

\section{Prompt for Summary Generation}
\label{sec:summary-prompt}

We use two distinct prompts for summary generation: (1) FeedSum Benchmark, where we generate summaries of varying quality using 13 different language models\footnote{Non-LLMs, such as Bart-large and Pegasus-large, do not require any prompt as they directly generate summaries from input documents. This is because they were trained via supervised fine-tuning on summarization datasets like CNNDM.} as summarizers, and (2) Evaluation of summarizers (i.e., Llama3 variants) after SFT or DPO. 

For the former, we use a simple prompt: 

\noindent{\sc Instruction: Summarize the Text.

\noindent Provide your answer in JSON format. The answer should be a dictionary with the key "summary" containing a generated summary as a string: \{"summary": "your summary"\}}.

For the latter, we use the exact same prompt to generate summaries across all Llama3 variants, identical to the input prompt shown in Table \ref{table:dpo-prompt-example}.

\section{Results with Gemma-2b-instruct}
\label{sec:exp-gemma}

We conduct an additional experiment to evaluate the improvements from preference learning with LLM-generated feedback, using a different LLM. Specifically, we choose Gemma-2b-instruct, as this smaller model demonstrates the impact of our framework, even when compared to significantly larger models like Llama-8/70b-instruct.

Table \ref{table:exp-dpo-gemma} presents the automated evaluation results across three fine-grained dimensions of summary quality, comparing their percentage scores before and after applying DPO with feedback from C4. LLM-generated feedback leads to significant improvements across all dimensions. However, these improvements are smaller than those observed in the Llama3 family (Table \ref{table:exp1}), indicating that larger LLMs benefit more from preference optimization with LLM-generated feedback in the context of text summarization.

Lastly, the significant performance gap between Gemma and Llama3 is primarily due to Gemma's inability to generate summaries for longer documents. When we test it, Gemma produces a weird response, e.g., {\sc The aim was a priori knowledge about about about about about about about about about about about about about about (...)}. This suggests that 2-billion-parameter models may not be suitable for long document summarization.

\section{Automated Evaluation using GPT-4o}
\label{sec:gpt-4o-eval}

\begin{table*}[t]
\begin{center}
\footnotesize
\begin{tabular}{ |X{2.2cm} |X{1.2cm} |X{1.0cm} X{1.2cm} X{2.1cm} |X{1.1cm} X{1.1cm} X{1.1cm} |X{1.1cm}|} \toprule
Backbone & \!\!Config.\!\! & Quality & \!\!Dimension\!\! & Granularity &  Faith. & Comp. & Conc. & Avg. \\ \midrule
{Llama3-8b-inst.} & \multicolumn{4}{c|}{wo. RL} &  0.864 & 0.526 & 0.439 & 0.610 \\ \midrule
\multirow{4}{*}{Llama3-8b-inst.} & DPO-C1 & Low & Single & Coarse-grained & 0.843 & 0.533 & 0.440 & 0.605 \\
 & DPO-C2 & High & Single & Coarse-grained & 0.880 & 0.554 & 0.519	 & 0.651 \\
& DPO-C3 & High & Multiple & Coarse-grained  & 0.881 & 0.530 & 0.519 & 0.643 \\
& DPO-C4 & High & Multiple & Fine-grained & \textbf{0.901} & \textbf{0.567} & \textbf{0.638} & \textbf{0.702} \\ \hline\midrule
Llama3-70b-inst.\!\! & \multicolumn{4}{c|}{wo. RL} & 0.925 & 0.554 & 0.484 & 0.654 \\ \midrule
Llama3-70b-inst.\!\! & DPO-C4 & High & Multiple & Fine-grained  & \textbf{0.934} & \textbf{0.581} & \textbf{0.738} & \textbf{0.751} \\\bottomrule
\end{tabular}
\end{center}
\vspace*{-0.4cm}
\caption{Automated evaluation results of seven summarizers on faithfulness, completeness, and conciseness using \textbf{GPT-4o as the FineSurE's automated evaluator}. The best score are marked in bold.} 
\label{table:exp1-gpt4o}
\vspace*{-0.45cm}
\end{table*}

Table \ref{table:exp1-gpt4o} summarizes the automated evaluation results using GPT-4o as the backbone for FineSurE.
While the numbers show a slight decrease compared to evaluations using Llama3-70B-Instruct in Table \ref{table:exp1}, the overall trends remain consistent. All observations (highlighted in bold in Section 5.1) remain valid even when using a different LLM (GPT-4o) as the fine-grained summary evaluator.

\section{Summary Example}
\label{sec:summ-example}

Table \ref{table:summ_example} presents examples of summaries generated by six different approaches: Llama3-8b/70b-instruct without DPO, and four variants of Llama3-8b-instruct after applying DPO.

The summary of DPO-C4 can be considered the best for the following reasons:

\smallskip
\noindent$\bullet$ \textbf{Core Focus:} The summary accurately captures the main theme of the conversation, which revolves around the Thanksgiving dinner arrangements. It highlights how the two people confirm plans, discuss what to bring, and finalize the decision for Person2 to bring wine instead of pie. This maintains the core context.

\smallskip
\noindent$\bullet$ \textbf{Inclusion of Key-facts:} The summary covers the important details of the conversation, including Person2's initial offer to bring dessert (pumpkin pie) and the shift to bringing wine due to another family member handling dessert. Other summaries tend to overlook or simplify this progression, while DPO-C4 fully captures the interaction’s key events.

\smallskip
\noindent$\bullet$ \textbf{Clarity and Conciseness:} The summary is structured in a straightforward, concise manner, effectively summarizing the conversation without unnecessary details. It presents the flow and outcome of the discussion clearly, making it easy for readers to understand. The logical order of events is maintained, ensuring a smooth narrative.

\smallskip
\noindent$\bullet$ \textbf{Accurate Role Depiction:} The summary clearly identifies Person1 as the host and Paul (Person2) as the guest, which helps clarify their relationship and the nature of the conversation. This distinction is more explicit in DPO-C4 compared to other summaries, which might leave these roles more ambiguous.

In conclusion, DPO-C4 is the best summary because it captures the essential points of the conversation with clarity and completeness, while maintaining a concise and well-structured form. It ensures that all significant details are included without overwhelming the reader.

\begin{table*}[t]
\begin{center}
\footnotesize
\begin{tabular}{|X{1.8cm} |L{13.3cm}|} \toprule
Faithfulness &  You will be given an article. You will then be given one summary written for this article. Your task is to rate the summary on one metric. Please make sure you read and understand these instructions carefully. Please keep this document open while reviewing, and refer to it as needed.

\vspace*{0.3cm}
Evaluation Criteria:
Consistency (1-5) - the factual alignment between the summary and the summarized source. A factually consistent summary contains only statements that are entailed by the source document. Annotators were also asked to penalize summaries that contained hallucinated facts. 

\vspace*{0.3cm}
Evaluation Steps:

1. Read the news article carefully and identify the main facts and details it presents.

2. Read the summary and compare it to the article. Check if the summary contains any factual errors that are not supported by the article.

3. Assign a score for consistency based on the Evaluation Criteria.

\vspace*{0.3cm}
Example:

\vspace*{0.3cm}
Source Text: 
{\color{blue}\{source text\}}

\vspace*{0.3cm}
Summary: 
{\color{blue}\{summary\}}


\vspace*{0.3cm}
Evaluation Form (scores ONLY):

\vspace*{0.3cm}
- Consistency: \\\midrule
Completeness &  You will be given an article. You will then be given one summary written for this article. Your task is to rate the summary on one metric. Please make sure you read and understand these instructions carefully. Please keep this document open while reviewing, and refer to it as needed.

\vspace*{0.3cm}
Evaluation Criteria:

Completeness (1-5) - the degree to which the summary includes all key information present in the source document. A complete summary accurately captures the main points, ideas, and relevant details without omitting crucial elements.

\vspace*{0.3cm}
Evaluation Steps:

1. Read the news article carefully and identify the main points, key information, and relevant details.

2. Read the summary and compare it to the article. Check if the summary captures all essential facts, main ideas, and pertinent details presented in the original article.

3. Assign a score for completeness based on the Evaluation Criteria.

\vspace*{0.3cm}
Example:

\vspace*{0.3cm}
Source Text: 
{\color{blue}\{source text\}}

\vspace*{0.3cm}
Summary: 
{\color{blue}\{summary\}}


\vspace*{0.3cm}
Evaluation Form (scores ONLY):

\vspace*{0.3cm}
- Completeness: \\\midrule
Conciseness &  You will be given an article. You will then be given one summary written for this article.
Your task is to rate the summary on one metric.
Please make sure you read and understand these instructions carefully.
Please keep this document open while reviewing, and refer to it as needed.

\vspace*{0.3cm}
Evaluation Criteria:

\vspace*{0.3cm}
Conciseness (1-5) - the extent to which the summary presents information succinctly and without unnecessary elaboration. A concise summary effectively conveys the essential content of the source document using clear and concise language, avoiding redundant or superfluous information.

\vspace*{0.3cm}
Evaluation Steps:

1. Read the news article carefully and identify the main points, key information, and relevant details.

2. Read the summary and compare it to the article. Check if the summary effectively conveys the essential content of the document in a concise manner, without unnecessary elaboration or redundancy.

3. Assign a score for conciseness based on the Evaluation Criteria.

\vspace*{0.3cm}
Example:

\vspace*{0.3cm}
Source Text: 
{\color{blue}\{source text\}}

\vspace*{0.3cm}
Summary: 
{\color{blue}\{summary\}}

\vspace*{0.3cm}
Evaluation Form (scores ONLY):

\vspace*{0.3cm}
- Conciseness: \\\bottomrule
\end{tabular}
\end{center}
\vspace*{-0.4cm}
\caption{Prompt of the modified G-Eval for faithfulness, completeness, and conciseness.}
\label{table:modified-geval}
\vspace*{-0.4cm}
\end{table*}

\begin{table*}[t]
\begin{center}
\footnotesize
\begin{tabular}{|X{1.8cm} |L{13.3cm}|} \toprule
Fact Check &  You will receive a document followed by a corresponding summary. 

Your task is to assess the factuality of each summary sentence across nine categories:

* no error: the statement aligns explicitly with the content of the document and is factually consistent with it.

* out-of-context error: the statement contains information not present in the document.

* entity error: the primary arguments (or their attributes) of the predicate are wrong.

* predicate error: the predicate in the summary statement is inconsistent with the document.

* circumstantial error: the additional information (like location or time) specifying the circumstance around a predicate is wrong.

* grammatical error: the grammar of the sentence is so wrong that it becomes meaningless.

* coreference error: a pronoun or reference with wrong or non-existing antecedent.

* linking error: error in how multiple statements are linked together in the discourse (for example temporal ordering or causal link).

* other error: the statement contains any factuality error which is not defined here.

\vspace*{0.2cm}
Instruction:

First, compare each summary sentence with the document.

Second, provide a single sentence explaining which factuality error the sentence has.

Third, answer the classified error category for each sentence in the summary.

\vspace*{0.2cm}
Provide your answer in JSON format. The answer should be a list of dictionaries whose keys are "sentence", "reason", and "category":

[{"sentence": "first sentence", "reason": "your reason", "category": "no error"}, {"sentence": "second sentence", "reason": "your reason", "category": "out-of-context error"}, {"sentence": "third sentence", "reason": "your reason", "category": "entity error"},]

\vspace*{0.2cm}
Document:

{\color{blue}\{document\}}

\vspace*{0.2cm}
Summary with {\color{blue}$\{\#$ of sentences$\}$} sentences:

{\color{blue}\{sentences\}}

\vspace*{0.2cm}
JSON Output: \\ \midrule
Key-fact Alignment & You will receive a summary and a set of key facts for the same document. 

Your task is to assess if each key fact is inferred from the summary.

\vspace*{0.2cm}
Instruction:

First, compare each key fact with the summary.

Second, check if the key fact is inferred from the summary and then response "Yes" or "No" for each key fact. If "Yes", specify the line number(s) of the summary sentence(s) relevant to each key fact. 

\vspace*{0.2cm}
Provide your answer in JSON format. The answer should be a list of dictionaries whose keys are "key fact", "response", and "line number":

[{"key fact": "first key fact", "response": "Yes", "line number": [1]}, {"key fact": "second key fact", "response": "No", "line number": []}, {"key fact": "third key fact", "response": "Yes", "line number": [1, 2, 3]}]

\vspace*{0.2cm}
Summary:

{\color{blue}\{summary\}}

\vspace*{0.2cm}
{\color{blue}$\{\#$ of key facts$\}$} key facts:

{\color{blue}\{key facts\}}

\vspace*{0.2cm}
JSON Output: \\\bottomrule
\end{tabular}
\end{center}
\vspace*{-0.4cm}
\caption{Prompt of the FineSurE for fact checking and key-fact alignment tasks.}
\label{table:finesure-prompt}
\vspace*{-0.4cm}
\end{table*}

\begin{table*}[t]
\begin{center}
\footnotesize
\begin{tabular}{|X{1.4cm} |L{13.7cm}|} \toprule
Input & <|begin\_of\_text|><|start\_header\_id|>user<|end\_header\_id|>
\vspace*{0.2cm}
Below is an instruction that describes a task. Write a response that appropriately completes the request.

\vspace*{0.2cm}
\#\#\#Instruction:

Please summarize the input document.

\vspace*{0.2cm}
\#\#\#Input:

MICHELE NORRIS, Host: And as NPR's Jackie Northam reports, Zardari isn't getting much relief while he's here.</s>JACKIE NORTHAM: Brian Katulis, a senior fellow at the Center for American Progress, says the attack sent a clear signal to the Obama administration that despite efforts to curb militancy, there's a serious undercurrent of extremism that has grown and metastasized in Pakistan.</s>BRIAN KATULIS: The concern of infiltration of those extremist voices in Pakistani's security services, a country that has anywhere from 70 to a hundred nuclear weapons, is the thing that keeps many people in Washington up at night.</s>JACKIE NORTHAM: Security and counterterrorism efforts are among the issues discussed today by Presidents Zardari and Obama before the Holbrooke service. But Katulis says Zardari isn't the best man for that discussion because he doesn't hold much sway.</s>BRIAN KATULIS: The meeting between President Obama and Zardari is an example of head-of-state diplomacy at its most complicated. Zardari, as the head of civilian government, really doesn't have as much power over those security issues as do Ashfaq Kayani, the head of the Pakistani military, or Ahmad Shuja Pasha who's the head of the Inter-Services Intelligence.</s>JACKIE NORTHAM: Jamie Metzl, the executive vice president of the Asia Society, says Pakistan needs to do better to justify U.S. support.</s>JAMIE METZL: We've spent \$20 billion in Pakistan since 9/11, huge amounts of military aid; lately, more civilian-focused aid, and the situation in Pakistan seems to have gone from bad to worse.</s>JACKIE NORTHAM: Kamran Bokhari, with the intelligence firm STRATFOR, says there are two schools of thought in Washington over how to deal with Pakistan. One is that Pakistan is playing a double game with Washington.</s>KAMRAN BOKHARI: This view says we need to be able to sustain the pressure on Pakistan, they can definitely do more, they're just not doing it. On the other hand, there are those who say Pakistan is already quite weakened. So if we demand more from the Pakistanis, what that means is that there is a good chance that it could further undermine stability within Pakistan.</s>JACKIE NORTHAM: Jackie Northam, NPR News, Washington.

\vspace*{0.2cm}
\#\#\#Response:<|eot\_id|>\\\midrule
Output (Chosen) &  <|start\_header\_id|>assistant<|end\_header\_id|>

NPR's Jackie Northam reports on concerns in Washington regarding extremism in Pakistan, highlighting the challenges faced by President Zardari in addressing security and counterterrorism with President Obama. Experts like Brian Katulis and Jamie Metzl criticize Pakistan's handling of militancy and question the effectiveness of U.S. aid, given the country's deteriorating situation. Kamran Bokhari of STRATFOR outlines the debate in Washington on how to approach Pakistan, balancing the need for pressure against the risk of destabilizing the country further.<|eot\_id|>\\\midrule
Output (Rejected) &  <|start\_header\_id|>assistant<|end\_header\_id|>

Pakistani President Asif Ali Zardari arrived in Washington for a meeting with President Obama, but has faced mounting pressure at home following the attack on the U.S. Embassy in Islamabad. The attack has raised concerns that extremist groups may have infiltrated Pakistan's security services, which oversee the country's nuclear arsenal. Zardari is seen as having little influence over security issues, with the military and intelligence agencies wielding more power. The meeting between the two presidents was focused on security and counterterrorism efforts in Pakistan, but experts warn that Pakistan's instability could continue to pose a danger to U.S. interests.<|eot\_id|> \\\bottomrule
\end{tabular}
\end{center}
\vspace*{-0.4cm}
\caption{An example of an input (user prompt) and its corresponding chosen and rejected outputs (assistant prompt) used to train Llama3-8b-instruct using DPO.}
\label{table:dpo-prompt-example}
\vspace*{-0.4cm}
\end{table*}

\begin{table*}[t]
\begin{center}
\footnotesize
\begin{tabular}{|X{2.2cm} |L{12.9cm}|} \toprule
Summarizer & \hspace*{5cm}FineSurE's Results \\ \toprule
GPT-4-turbo & \textbf{Task 1. Fact Checking}

[
\{"sentence": "NPR's Jackie Northam reports on concerns in Washington regarding extremism in Pakistan, highlighting the challenges faced by President Zardari in addressing security and counterterrorism with President Obama.", "reason": "The sentence accurately summarizes the content of the document.", "category": "no error"\},

\{"sentence": "Experts like Brian Katulis and Jamie Metzl criticize Pakistan's handling of militancy and question the effectiveness of U.S. aid, given the country's deteriorating situation.", "reason": "The sentence accurately summarizes the opinions of Brian Katulis and Jamie Metzl in the document.", "category": "no error"\},

\{"sentence": "Kamran Bokhari of STRATFOR outlines the debate in Washington on how to approach Pakistan, balancing the need for pressure against the risk of destabilizing the country further.", "reason": "The sentence accurately summarizes Kamran Bokhari's views on the debate in Washington.", "category": "no error"\}
]

\vspace*{0.3cm}

\textbf{Faithfulness Score = 3/3 = 100\%}

\vspace*{0.3cm}

\textbf{Task 2. Key-fact Alignment}

[
\{"key fact": "Pakistani President Asif Ali Zardari is visiting Washington this week", "response": "Yes", "line number": [1]\},

\{"key fact": "Asif Ali Zardari is attending a memorial for Richard Holbrooke", "response": "No", "line number": []\},

\{"key fact": "Officials describe Zardari's visit as private", "response": "No", "line number": []\},

\{"key fact": "Pakistan is currently experiencing political turmoil", "response": "Yes", "line number": [2, 3]\},

\{"key fact": "The U.S. is pressuring Pakistan to curb terrorism", "response": "Yes", "line number": [1, 3]\}
]

\vspace*{0.3cm}

\textbf{Completeness Score = 3/5 = 60\%}

\textbf{Conciseness Score = 3/3 = 100\%}
\\ \midrule
Mistral-7b-inst & \textbf{Task 1. Fact Checking}

[
\{"sentence": "Pakistani President Asif Ali Zardari arrived in Washington for a meeting with President Obama, but has faced mounting pressure at home following the attack on the U.S. Embassy in Islamabad.", "reason": "The document does not mention the attack on the U.S. Embassy in Islamabad.", "category": "out-of-context error"\},

\{"sentence": "The attack has raised concerns that extremist groups may have infiltrated Pakistan's security services, which oversee the country's nuclear arsenal.", "reason": "The document mentions concerns about infiltration of extremist voices in Pakistani's security services, but not specifically the attack raising these concerns.", "category": "out-of-context error"\},

\{"sentence": "Zardari is seen as having little influence over security issues, with the military and intelligence agencies wielding more power.", "reason": "The document explicitly states that Zardari doesn't hold much sway over security issues, and that the head of the Pakistani military and the head of the Inter-Services Intelligence have more power.", "category": "no error"\},

\{"sentence": "The meeting between the two presidents was focused on security and counterterrorism efforts in Pakistan, but experts warn that Pakistan's instability could continue to pose a danger to U.S. interests.", "reason": "The document mentions that security and counterterrorism efforts were discussed, but does not mention experts warning about Pakistan's instability posing a danger to U.S. interests.", "category": "out-of-context error"\}
]

\vspace*{0.3cm}

\textbf{Faithfulness Score = 1/4 = 25\%}

\vspace*{0.3cm}

\textbf{Task 2. Key-fact Alignment}

[
\{"key fact": "Pakistani President Asif Ali Zardari is visiting Washington this week", "response": "Yes", "line number": [1]\},

\{"key fact": "Asif Ali Zardari is attending a memorial for Richard Holbrooke", "response": "No", "line number": []\},

\{"key fact": "Officials describe Zardari's visit as private", "response": "No", "line number": []\},

\{"key fact": "Pakistan is currently experiencing political turmoil", "response": "Yes", "line number": [1]\},

\{"key fact": "The U.S. is pressuring Pakistan to curb terrorism", "response": "Yes", "line number": [4]\}
]

\vspace*{0.3cm}

\textbf{Completeness Score = 3/5 = 60\%}

\textbf{Conciseness Score = 2/4 = 50\%} \\\bottomrule
\end{tabular}
\end{center}
\vspace*{-0.4cm}
\caption{Results by FineSurE\,\cite{song2024finesure} on the "chosen" and "rejected" summaries in Table \ref{table:dpo-prompt-example}.}
\label{table:ex-finesure}
\vspace*{-0.4cm}
\end{table*}

\begin{table*}[t]
\begin{center}
\footnotesize
\begin{tabular}{|X{2.2cm} |L{12.9cm}|} \toprule
Document & 
\#Person1\#: Hey, Paul, you're still having Thanksgiving dinner at my house on Thursday, right?

\#Person2\#: Yeah, thanks again for the invitation. I was worried I'd have to spend it alone after my parents announced they'd be in Finland.

\#Person1\#: I think it's strange that you're not going with them. Is it because of the cold weather in Northern Europe?

\#Person2\#: Not exactly. I have a lot of work to do next week.

\#Person1\#: Well, I'm glad you won't be alone on a holiday. It'll be great to have you celebrate with us. The more the merrier!

\#Person2\#: Exactly, how many people will be there?

\#Person1\#: Let's see, my parents, my grandpa, my sister, my brother-in-law, my niece and my two cousins, plus us. So, ten people altogether.

\#Person2\#: Should I bring anything?

\#Person1\#: Nope, we've got it covered.

\#Person2\#: Come on, I hate to show up empty-handed. You are being so generous to have me. I know! I'll bring pie. I'll make a great pumpkin pie.

\#Person1\#: Sounds delicious, but my sister has already volunteered for dessert. Look, don't worry about it. There will be tons of food. Turkey, stuffing, green beans, sweet potatoes.

\#Person2\#: Wow, I'm getting hungry just thinking about it. How about this, I'll bring a nice bottle of wine. You can never have too much of that.

\#Person1\#: Well, isn't that the truth? Sounds good to me.
\\\midrule
\makecell{Llama3-8b-inst.\\(wo. RL)} & Person1 invites Person2 to join them for Thanksgiving dinner at their house. Person2 accepts the invitation, citing that they were worried they would have to spend the holiday alone after their parents announced they would be in Finland. Person1 reassures Person2 that they are welcome and that the more the merrier. Person2 asks how many people will be attending and is told that there will be 10 people, including their family and cousins. Person2 offers to bring a gift, but Person1 declines, saying that they have everything covered. Person2 persists and offers to bring a bottle of wine or a pie, but Person1 suggests that their sister has already volunteered to bring dessert.
\\\midrule
\makecell{\!\!\!\!Llama3-70b-inst.\!\!\!\!\\(wo. RL)} & Person 1 invites Person 2 to their house for Thanksgiving dinner on Thursday. Person 2 is grateful for the invitation and explains they can't go to Finland with their parents due to work commitments. Person 1 is happy to have them celebrate with their family, which will consist of 10 people. When Person 2 offers to bring something, Person 1 declines, but eventually agrees to let them bring a bottle of wine.
\\\midrule
\makecell{\!\!Llama3-8b-inst.\!\!\\(w. DPO-C1)} & Paul and Person2 confirm their plans for Thanksgiving dinner at Person1's house, discussing the guest list and what each person will bring. Person2 offers to bring a pumpkin pie, but Person1 declines, mentioning that someone else is already providing dessert. Person2 then offers to bring a bottle of wine instead, which Person1 accepts.
\\\midrule
\makecell{\!\!Llama3-8b-inst.\!\!\\(w. DPO-C2)}  & Person1 and Person2 are confirming their plans for Thanksgiving dinner at Person1's house on Thursday. Person2 is grateful for the invitation and offers to bring a contribution, initially suggesting pie, but then changes to a bottle of wine instead, since Person1's sister is already handling dessert. Person1 is appreciative of the offer, but insists that they have everything covered and that the wine sounds like a great addition.
\\\midrule
\makecell{\!\!Llama3-8b-inst.\!\!\\(w. DPO-C3)}  & Person1 invites Person2 to join their family's Thanksgiving dinner, which Person2 had initially worried they might spend alone due to their parents traveling to Finland. Person2 expresses gratitude and offers to bring something, but Person1 assures them that they have everything covered. Person2 decides to bring a bottle of wine instead, which is accepted with enthusiasm.
\\\midrule
\makecell{\!\!Llama3-8b-inst.\!\!\\(w. DPO-C4)}  & The conversation is about Thanksgiving dinner arrangements, with Person2 (Paul) confirming his attendance at Person1's (host's) house on Thursday, discussing what to bring and what's already been taken care of, and finalizing the offer to bring wine instead of pie since someone else is handling desserts.
\\\bottomrule
\end{tabular}
\end{center}
\vspace*{-0.4cm}
\caption{Example of summaries for an identical dialogue from DialogSum.}
\label{table:summ_example}
\vspace*{-0.4cm}
\end{table*}

\end{document}